\documentclass[twoside]{article}

\usepackage{ecj,palatino,epsfig,latexsym,natbib}

\usepackage{multirow}

\usepackage{algorithm}
\usepackage{algpseudocode}
\algnewcommand\algorithmicforeach{\textbf{for each}}
\algdef{S}[FOR]{ForEach}[1]{\algorithmicforeach\ #1\ \algorithmicdo}
\usepackage{algpseudocode}
\algrenewcommand\textproc{}

\let\oldReturn\Return
\renewcommand{\Return}{\State\oldReturn}
\usepackage{comment}

\usepackage{url}
\usepackage{lscape}
\usepackage{amsmath,amssymb,amsfonts}
\usepackage{graphicx}
\usepackage{textcomp}
\usepackage{xcolor}
\usepackage{color,soul}
\usepackage{colortbl}
\def\BibTeX{{\rm B\kern-.05em{\sc i\kern-.025em b}\kern-.08em
  T\kern-.1667em\lower.7ex\hbox{E}\kern-.125emX}}

\usepackage[caption=false]{subfig}
\graphicspath{{./figs/}}

\makeatletter
\newenvironment{subfigures}
 {\begin{minipage}{\columnwidth}\def\@captype{figure}\centering}
 {\end{minipage}}
\makeatother

\usepackage{tikz}
\tikzset{node_style/.style={circle,draw=black,fill=gray!20!}}
\tikzset{edge_style/.style={draw=black, thick}}
\tikzset{gexamples_node_style/.style={circle,draw=black,fill=gray!20!,scale=0.8}}

\begin{document}

\ecjHeader{x}{x}{xxx-xxx}{2020}{Evolving Plasticity under Changing Environmental Conditions}{A. Yaman et al.}
\title{\bf Evolving Plasticity for Autonomous Learning\\ under Changing Environmental Conditions}  

\author{\name{\bf Anil Yaman} \hfill \addr{anilyaman@kaist.ac.kr}\\ 
        \addr{Department of Mathematics and Computer Science, Eindhoven University of Technology, Eindhoven, 5612 AP, the Netherlands\\ Department of Bio and Brain Engineering, Korea Advanced Institute of Science and Technology, Daejeon, 34141, Republic of Korea
        }
\AND
       \name{\bf Giovanni Iacca} \hfill \addr{giovanni.iacca@unitn.it}\\
        \addr{Department of Information Engineering and Computer Science, University of Trento, Trento, 38122, Italy
        }
\AND
       \name{\bf Decebal Constantin Mocanu} \hfill \addr{d.c.mocanu@utwente.nl}\\
        \addr{Department of Mathematics and Computer Science, Eindhoven University of Technology, Eindhoven, 5612 AP, the Netherlands\\
        Faculty of Electrical Engineering, Mathematics and Computer Science, University of Twente, Enschede, 7522NB, the Netherlands
        }
\AND
       \name{\bf Matt Coler} \hfill \addr{m.coler@rug.nl}\\
        \addr{Campus Frysl\^an, University of Groningen, Leeuwarden, 8911 AE, the Netherlands
        }
\AND
       \name{\bf George Fletcher} \hfill \addr{g.h.l.fletcher@tue.nl}\\
        \addr{Department of Mathematics and Computer Science, Eindhoven University of Technology, Eindhoven, 5612 AP, the Netherlands
        }
\AND
       \name{\bf Mykola Pechenizkiy} \hfill \addr{m.pechenizkiy@tue.nl}\\
        \addr{Department of Mathematics and Computer Science, Eindhoven University of Technology, Eindhoven, 5612 AP, the Netherlands}
}

\maketitle

\begin{abstract}
A fundamental aspect of learning in biological neural networks is the plasticity property which allows them to modify their configurations during their lifetime. Hebbian learning is a biologically plausible mechanism for modeling the plasticity property in artificial neural networks (ANNs), based on the local interactions of neurons. However, the emergence of a coherent global learning behavior from local Hebbian plasticity rules is not very well understood. The goal of this work is to discover interpretable local Hebbian learning rules that can provide autonomous global learning. To achieve this, we use a discrete representation to encode the learning rules in a finite search space. These rules are then used to perform synaptic changes, based on the local interactions of the neurons. We employ genetic algorithms to optimize these rules to allow learning on two separate tasks (a foraging and a prey-predator scenario) in online lifetime learning settings. The resulting evolved rules converged into a set of well-defined interpretable types, that are thoroughly discussed. Notably, the performance of these rules, while adapting the ANNs during the learning tasks, is comparable to that of offline learning methods such as hill climbing.

\end{abstract}


\section{Introduction}
\label{sec:intro}
In the past few decades, a broad area of research in nature-inspired hardware and software design~\citep{de2006fundamentals, sipper1997phylogenetic} has been stimulated by the study of the evolutionary, developmental and learning processes that allowed biological organisms to adapt to their environment. In particular, artificial neural networks (ANNs) have proved to be a successful -yet simplified- formalization of the information processing capability of biological neural networks (BNNs)~\citep{rumelhart1986learning}. 

Inspired by the evolutionary process of biological systems, the research field known as \textit{Neuroevolution} (NE) employs evolutionary computing (EC) approaches to optimize ANNs~\citep{floreano2008neuroevolution, yao1999evolving}. Adopting the terminology from biology, a population of individuals are represented as individual \textit{genotypes}, which encode the parameters (i.e. topology, weights and/or the learning approaches) of the ANNs. Biologically inspired operators, namely \textit{selection}, \textit{crossover} and \textit{mutation}, are iteratively applied to generate new individuals and find ANNs that are better adapted to the task at hand~\citep{Goldberg1989}.

One key aspect in NE is the encoding of the ANNs. This, in turn, influences the so-called genotypes-phenotype mapping, i.e. the way a given genotype is used to build a certain phenotype. In \textit{direct encoding}, the parameters of the networks (mainly weights and/or topology) are directly encoded into the genotype of the individuals and optimized to solve the task~\citep{yaman2018limited, mocanu2018scalable}. However, these parameters usually remain fixed during the network's lifetime, such that it cannot adapt if the environment changes. In \textit{indirect encoding}, on the other hand, some kinds of rules for the network's development and/or training are optimized~\citep{mouret2014artificial, nolfi1994phenotypic}. For instance, the plasticity property observed in BNNs has been modelled in various works \citep{yaman2019dsp, soltoggio2017born, soltoggio2008evolutionary, kow2016growing} to obtain evolving plastic artificial neural networks (EPANNs). 

The traditional plasticity model is based on a biologically plausible mechanism known as \textit{Hebbian learning} \citep{hebb1949, kuriscak2015}, which performs synaptic adjustments using plasticity rules based on the local activations of neurons. One important limitation of the basic form of Hebbian learning is its instability, due to possibly indefinite increase/decrease of the synaptic weights~\citep{vasilkoski2011review}. To overcome this limitation, several variants of Hebbian learning have been proposed which stabilize the learning process in various ways~\citep{brown1990hebbian, vasilkoski2011review, sejnowski1989hebb}. Nevertheless, these improved plasticity rules may still require further optimization to properly capture the dynamics needed for adjusting the network parameters to a given task.

A number of works used EC to optimize plasticity in ANNs to achieve lifetime learning~\citep{coleman2012evolving}. Some of these works optimize the parameters of the Hebbian learning rules~\citep{floreano2000evolutionary, niv2002evolution}. Some other works replace the Hebbian rules with evolving complex functions, such as using an additional ANN which determines the synaptic changes~\citep{orchard2016evolution, risi2010indirectly}. Others use a mechanism known as \textit{neuromodulation}, in which the ANNs include specialized neurons that are used to signal the synaptic changes between other neurons~\citep{runarsson2000evolution, soltoggio2008evolutionary}.

Despite the relatively vast literature on the use of EC to discover and optimize plasticity models, we believe that those previous models are either too simple to obtain actual adaptation, or too complex to \textit{understand} the evolved learning behavior. In fact, although complex plasticity models can provide a solution to certain learning problems, their complexity may prevent gaining insights into the learning behavior of the networks\footnote{This is the case of some of the existing works where the initial synaptic weights and/or the connectivity of the networks is evolved in addition to the plasticity rules~\citep{orchard2016evolution,soltoggio2008evolutionary}. As for the initial synaptic weights, evolving them obviously increases the number of parameters. Furthermore, this aspect can be decoupled from the evolution of plasticity rules \textit{per se}. Likewise, evolving the connectivity of the networks may overfit the networks to a certain task, making it difficult to evolve an actually adaptive behavior.}.

In this work, we use binary neuron activations to reduce the number of pairwise activation states of neurons, and evolve discrete plasticity rules to perform synaptic changes based on pairwise neuron activation states and reinforcement signals. An important advantage of our model is that, due to the discrete nature of these rules, it is possible to \textit{interpret} the learning behavior of the networks. We use networks consisting of one hidden layer, where we introduce local weight competition to allow self-organized adaptation of synaptic weights. We demonstrate the lifetime learning and adaptation capabilities of plastic ANNs on two separate tasks, namely a foraging and a prey-predator scenario. In both scenarios, starting from a randomly initialized ANN an agent is required to learn to perform certain actions during its lifetime, in a continuous learning setting. We then show that: (1) the evolved synaptic plasticity (ESP) rules we obtain with our model are capable of producing stable learning and adaptation capabilities; (2) the ESP rules are intelligible as they can be easily interpreted and linked to the task at hand\footnote{As it will become clear in Section~\ref{sec:experimentalSetup}, we distinguish two levels of complexity, which are intertwined: the first level concerns the task, the second one concerns the optimization of the ANNs using the ESP rules. Even in simpler tasks, optimizing the parameters of the ANNs could be challenging for plasticity rules, since the process is affected by the size of the network: indeed, larger networks involve a large number of parameters and pairwise interactions, which can make it harder to optimize.}. Finally, we compare our results with the Hill Climbing (HC) algorithm~\citep{de2006fundamentals}.

The rest of this paper is organized as follows: in Section~\ref{sec:methods}, we provide background knowledge on Hebbian learning and introduce our approach to represent and evolve Hebbian based synaptic plasticity rules; in Section~\ref{sec:experimentalSetup}, we discuss our experimental setup, and introduce the foraging and pre-predator tasks; in Section~\ref{sec:results}, we provide the results of the proposed approach and compare it with the Hill Climbing algorithm; finally, in Section~\ref{sec:conclusion} we summarize our conclusions and discuss future work.

\section{Evolution of Plasticity Rules}
\label{sec:methods}
In its general form, Hebbian learning is formulated as:
\begin{equation}
w_{i,j}{(t+1)} = w_{i,j}{(t)} + \Delta w_{i,j}
\label{eq:hebbianLearning}
\end{equation}
\begin{equation}
\Delta w_{i,j} = f(a_i,a_j,m) = \eta  a_i a_j m \label{eq:plainHebbianRule}
\end{equation}
where the synaptic efficiency\footnote{In the following, we will use ``efficiency'' and ``weight'' interchangeably.} $w_{i,j}$ at time $t+1$ is updated by the change $\Delta w_{i,j}$, that is a function of pre- and post-synaptic activations, $a_i$ and $a_j$, and a modulatory signal, $m$. This function is considered the same for all the weights in the network, i.e. $f(a_i,a_j,m)$ does not depend on $i$ and $j$. 

The plain Hebbian rule, given in Equation~\ref{eq:plainHebbianRule}, strengthens the synaptic efficiency $w_{i,j}$ when the signs of $a_i$ and $a_j$ are positively correlated, weakens it when the signs are negatively correlated, and keeps it stable when at least one of the two activations is zero~\citep{brown1990hebbian}. The modulatory signal, $m$, is used to determine the sign of the Hebbian rule. When $m$ is positive, plain Hebbian learning is performed. However, when $m$ is negative, the sign of the learning rule is reversed (this is also known as \textit{anti-Hebbian learning}), i.e. the synaptic efficiency is strengthened if the activations of neurons are negatively correlated, and weakened if they are positively correlated. Additionally, a constant $\eta$ is used as a learning rate to scale the magnitude of the synaptic change.

In this work, we use the step function (see Appendix~\ref{appx:anns}) to binarize the neuron activations, i.e. we apply Equations ~\ref{eq:hebbianLearning}-\ref{eq:plainHebbianRule} with binary values for $a_i$ and $a_j$. In addition to this, we set the modulatory signal, $m$, as:
\begin{equation}
m = \left\{
 \begin{array}{lr}
  +1, & \mbox{if desired output (reward);}\\
  -1, & \mbox{if undesired output (punishment);}\\
  0, & \mbox{otherwise (neutral).}
 \end{array}
\right. 
\label{eq:modulation}
\end{equation}
In other words, we use $m$ as a \emph{reinforcement signal} to indicate how the network is performing with respect to the task. Therefore, $m=1$ ($m=-1$) indicates that the current configuration of the network is producing a desired (undesired) behavior and Hebbian (anti-Hebbian) learning should be used to increase (reduce) the synaptic weights and promote (avoid) producing this behavior in the same situation. The desired and undesired outcomes are defined by a reward function which depends on the task, according to the desired/undesired associations between sensory inputs and behavioral outputs. The reward functions we used in our experiments are discussed in Section~\ref{sec:experimentalSetup}. Furthermore, for each neuron $i$ we scale the weights after each synaptic change (i.e., after Equations ~\ref{eq:hebbianLearning}-\ref{eq:plainHebbianRule}), as follows:
\begin{equation}
w^{\prime}_{i,j}(t+1) = \frac{w_{i,j}(t+1)}{|| \boldsymbol{w_{i}}(t+1) ||_2}
\label{eq:normEquation}
\end{equation}
where the vector $\boldsymbol{w_{i}(t+1)}$ encodes all the incoming weights to a post-synaptic neuron $i$. This normalization process prevents indefinite synaptic growth, and helps connections to specialize by introducing local synaptic competition, a concept observed also in biological neural networks~\citep{el2018locally}.
%This scaling allows to have a unit length using the Euclidean norm $|| \cdot||_2$.

\begin{figure}[!ht]
\begin{minipage}[c]{0.55\textwidth}
\includegraphics[width = \textwidth]{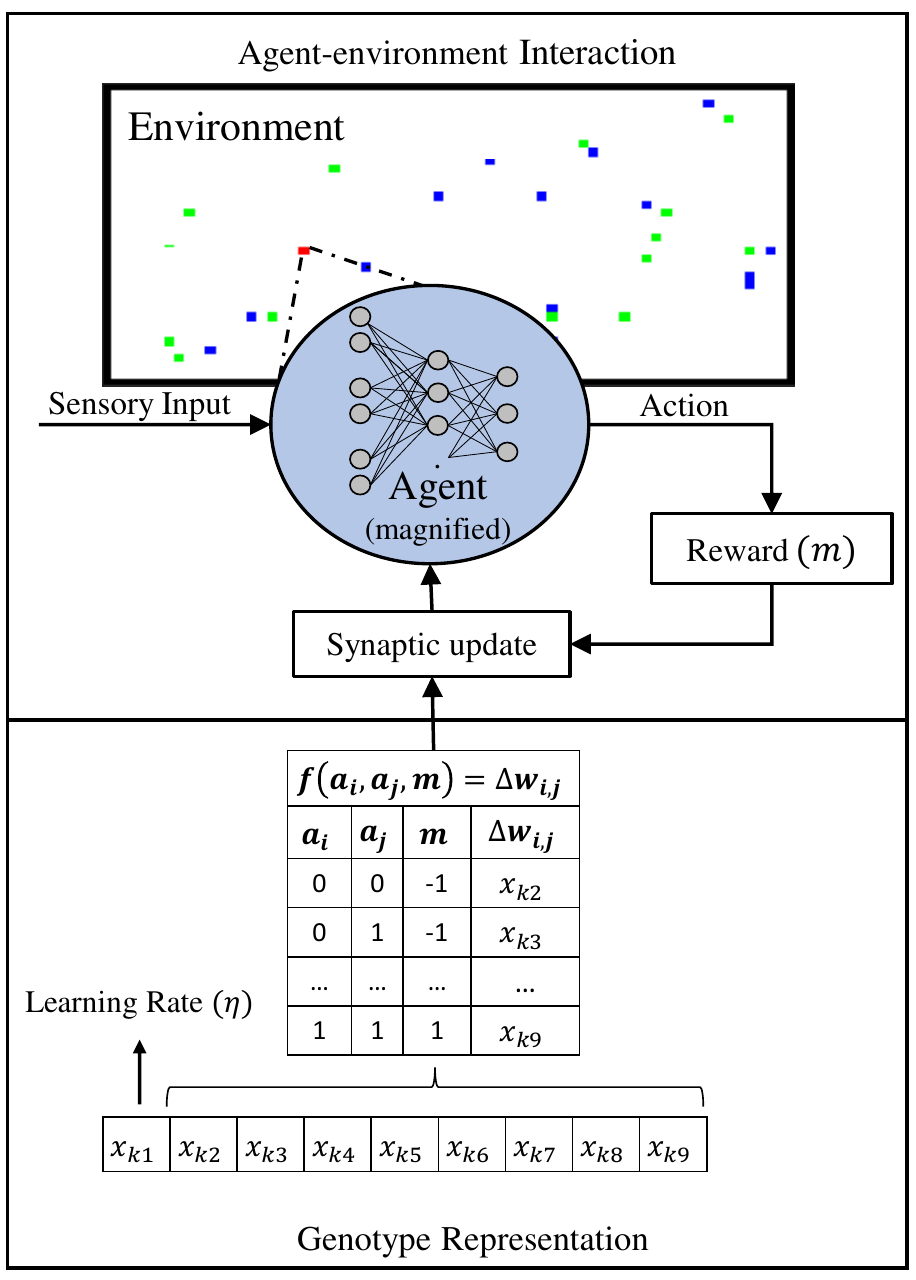}
\end{minipage}
\hfill
\begin{minipage}[c]{0.43\textwidth}
%\vspace{-3.8cm}
\caption{Genotype representation and agent-environment interaction. The genotypes of the individuals encode the learning rate and the synaptic update outcomes for $8$ possible states of $a_i$, $a_j$ and $m$. The agent (depicted in red) interacts, for a given number of steps, with a (task-specific) environment. An artificial neural network is used to control the actions of the agent. The initial weights of the ANN are randomly initialized, and are changed after each action step based on the ESP rules and a reinforcement signal received from the environment.}
\label{fig:framework}
\end{minipage}
\vspace{-0.5cm}
\end{figure}

To optimize the plasticity rules $\Delta w_{i,j} = f(a_i,a_j,m)$, we employ a Genetic Algorithm (GA)~\citep{Goldberg1989}. Figure~\ref{fig:framework} illustrates the genotype of the individuals, where we encode the learning rate $\eta\in [0,1)$, and one of three possible outcomes $\{-1,0,1\}$ (corresponding to \textit{decrease}, \textit{stable}, and \textit{increase}, respectively) for each plasticity rule. Since we use binary activations for neurons, there are $4$ possible activation combinations of $a_i$ and $a_j$. Furthermore, we take into account only positive and negative reinforcement signals, ignoring the case $m=0$ since the synaptic change is performed only when $m=+1$ or $m=-1$. Consequently, there are $8$ possible input combinations, and therefore $8$ possible plasticity rules defined by $f(a_i,a_j,m)$. The size of the search space of the plasticity rules is then $3^8$, excluding the real-valued learning rate $\eta$.

In the initialization step, we randomly initialize a population of $9$-dimensional individual vectors $\boldsymbol{x}_k$, where $k=(1,\hdots,N)$ (in our experiments, $N$ was set to $30$) to encode the synaptic update rules. Each dimension of the individuals is uniformly sampled within its domain, depending on its data type (real-valued, or discrete).

The evaluation process of an individual starts with the initialization of an agent with a random ANN configuration. Here, we use fixed topology fully connected feed-forward neural networks, with connection weights sampled (as real values) from a uniform distribution in $[-1,1]$. The agent is allowed to interact with the environment by performing actions based on the output of its controlling ANN. After each action step, the weights of the ANN are changed based on a synaptic update table. This table is constructed by converting the vector representation of the individual plasticity rules to specify how synaptic weights are modified based on the pre-, post-synaptic, and reinforcement signals (see Figure~\ref{fig:framework}). The evaluation process of the GA is based on a continuous learning setting, where the weights of the ANNs are updated constantly. 

We use an elitist roulette wheel selection operator to determine the individuals for the next generation: the top $10$ best individuals are copied to the next generation without any change; the remaining individuals are generated from fitness-proportionate selected parents using uniform crossover with a probability of $0.5$. After crossover, we perturb these individuals by applying a Gaussian mutation $\mathcal{N}(0,0.1)$ to the real-valued component $\eta$, with probability $1.0$, and re-sampling the $8$ discrete components, with a probability of $0.15$. The evolutionary process is executed until there is no more improvement in terms of best fitness for 30 generations.

\section{Experimental Setup}
\label{sec:experimentalSetup}
We test the learning and adaptation capabilities of the plastic ANNs with ESP rules on two agent-based tasks within reinforcement learning settings: a foraging and a prey-predator scenario. We discuss the specifics of these tasks in the following sections.

\subsection{Foraging Task}
\label{sec:foragingEnvironment}
In the foraging task, inspired by \cite{soltoggio2012modulated}, an artificial agent is required to learn to navigate within an enclosed environment and collect/avoid the correct types of items in the environment. 

\begin{figure*}[!ht]
\begin{subfigures}
\subfloat[Foraging environment]{\includegraphics[width=0.5\columnwidth]{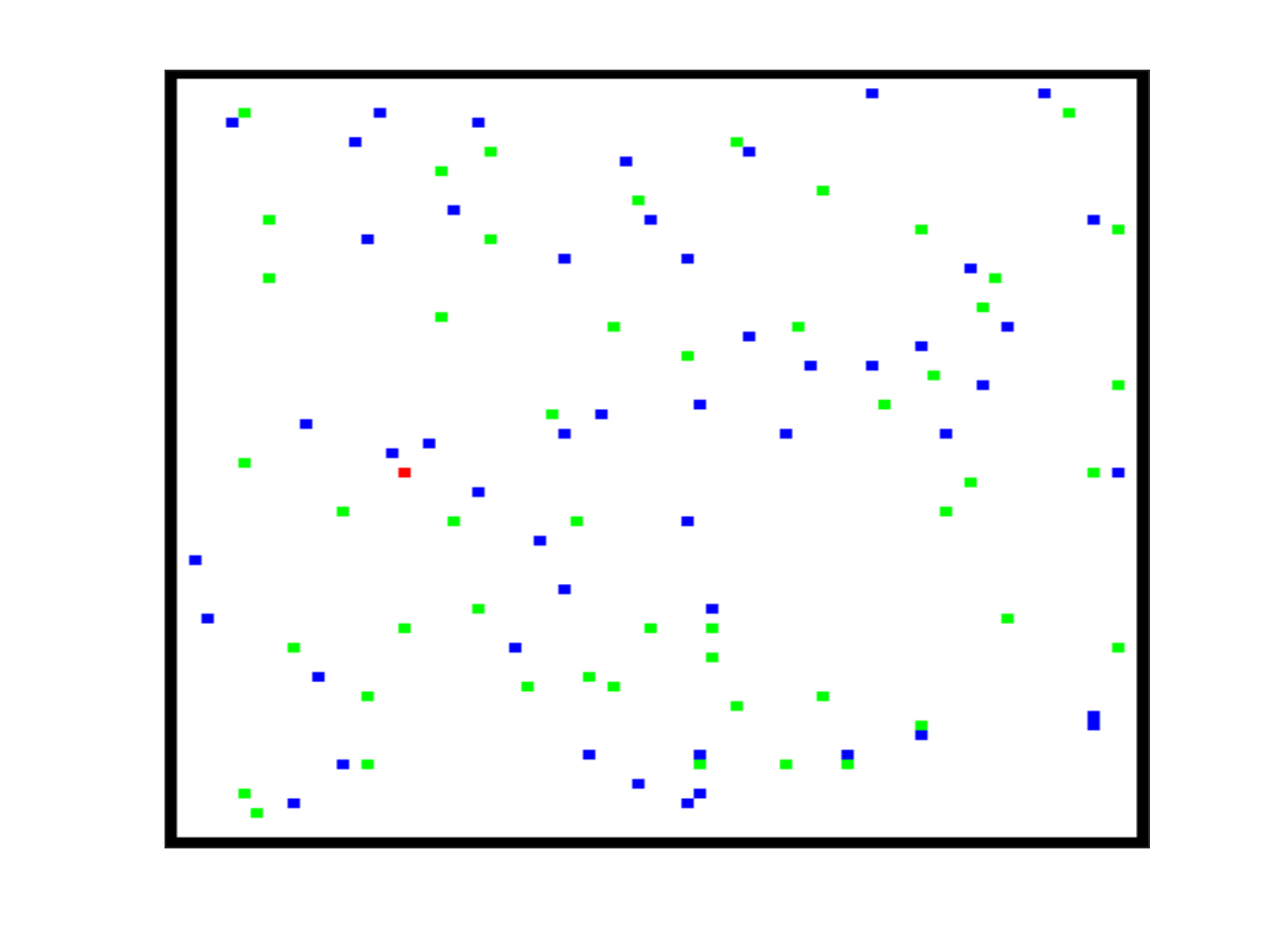}\label{fig:environment}}
\subfloat[Agent architecture]{\includegraphics[width=0.5\columnwidth]{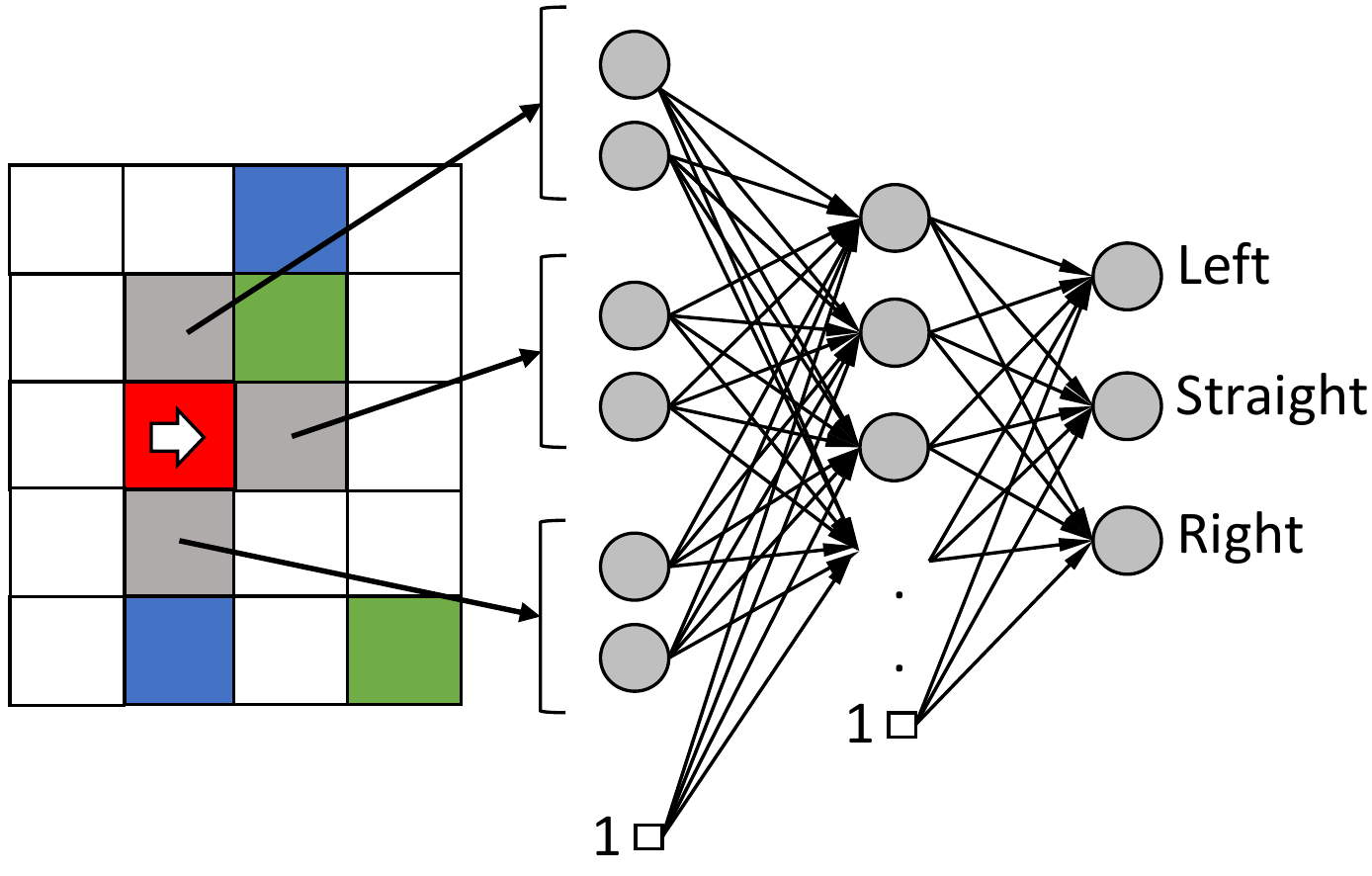}\label{fig:agentANN}}
\end{subfigures}
\caption{(a) Simulation environment used in the foraging task experiments. The location of the agent, two types of items and the wall are depicted in red, green, blue and black respectively. (b) Agent's position/direction (in red) and sensory inputs from left, front and right cells (in gray) within a foraging environment, and the ANN controller, with $1$ hidden layer, $6$ inputs ($2$ per cell), and $3$ outputs (Left/Straight/Right).}\label{fig:environmentAgent}
%\vspace{-0.5cm}
\end{figure*}

A visualization of the simulated environment is provided in Figure~\ref{fig:environment}. The environment consists of $100 \times 100$ grid cells enclosed by a wall. The agent, shown in red, has a direction to indicate its orientation on the grid. Two types of food items ($50$ green and $50$ blue) are randomly placed to fixed locations on the grid. The agent has sensors that can take inputs from the nearest cell on the left, in front, and on the right.

There are four possible states for each cell (empty, wall, green, blue), therefore we represent the sensor reading of each cell with two bits, as $[(0,0),(1,1),(1,0),(0,1)]$. The agent performs one of three possible actions (``Left'', ``Straight'', ``Right'') based on the output of its ANN. The output neuron with the maximum activation value is selected to be the action of the agent. In the cases of ``Left'' and ``Right'' the agent's direction is changed accordingly, and the agent is moved one cell in the corresponding direction. In the case of ``Straight'', the direction of the agent is kept constant and the agent is moved one cell along its original direction.

To test the adaptation abilities of the networks, we define two reward functions that we refer to as two ``seasons'': \textit{summer} and \textit{winter} (see Appendix~\ref{appx:sub:reinforcementFunction:foraging} for the complete set of associations between behaviors and reinforcement signals). In both seasons, the agent is expected to explore the environment and avoid collisions with the walls. During the summer season, the agent is expected to learn to collect green items and avoid blue items, while during the winter season this requirement is reversed. We calculate the average of the agent's performance score, per each season, by subtracting the number of incorrectly collected items from the number of correctly collected items. We start an experiment from the summer season. Then we switch season every $5000$ action steps, for a total of four seasons (summer, winter, summer, winter). It is important to note that these seasonal changes only cause the reinforcement associations to change, but they do not effect the network configuration. We perform this process for five independent trials, with each trial starting with a randomly initialized ANN. The fitness value of an individual plasticity rule is then given by:
\begin{equation}
\vspace{-0.1cm}
fitness =\frac{1}{S\cdot T} \sum_{k=1}^{S} \sum_{l=1}^{T}(c_{k,l} - i_{k,l})
\label{eq:fitness}
%\vspace{-0.1cm}
\end{equation}
where $T=5$ and $S=4$ are the number of trials and seasons, respectively, and $c_{k,l}$ and $i_{k,l}$ are the number of correctly and incorrectly collected items in each season $k$ of each trial $l$. When an item is collected, a new item of the same type is randomly placed on an unoccupied cell.

The architecture of the agent is illustrated in Figure~\ref{fig:agentANN}. We use fully connected feed-forward networks with one hidden layer, for a total of $6$ input, $20$ hidden and $3$ output neurons\footnote{We also performed additional experiments with networks consisting of various numbers of hidden neurons, see Appendix~\ref{appx:sub:SensitivityNumHiddenNeeurons} for details.}. One additional bias neuron is added to the input and hidden layers.

For all experiments, the agents are set to pick a random action with a probability of $0.02$, regardless of the actual output of their ANNs. Random behaviors are introduced to avoid getting stuck in a behavioral cycle. Such random behaviors, though, are not taken into account for the synaptic changes because they are not based on the output of the network.

%%%%%%%%%%%%%%%%%%%%%%%%%%%%%%%%%%%%%%%%%%%%%%%%%%%%%%%%%%%%%%%%%%%%

\subsection{Prey-predator Task}
\label{sec:preyPredator}
The prey-predator task includes two types of agents, referred to as preys and predators, that try to avoid and catch, respectively, the agent controlled by the ANN. The agent, starting from a randomly initialized ANN, is required to learn to catch preys and escape from predators during its lifetime, based on the ESP rules and the reinforcement signals.

Also in this case, the environment consists of $100 \times 100$ grid cells enclosed by a wall. We introduce $10$ mobile preys and $10$ predators, both controlled by hand-coded rules. In each action step, they all move to a randomly selected neighboring cell. When the agent is in close proximity (determined by a certain threshold based on Euclidean distance), preys move to a randomly selected neighboring cell with a higher probability of maximizing their distance to the agent; on the contrary, predators move to a randomly selected neighboring cell with a higher probability of minimizing their distances to the agent. If the agent moves to the same cell occupied by a prey, we randomly relocate the prey to another unoccupied cell and count this event as a ``collected" point for the agent. If a predator moves to the same cell occupied by the agent, we keep the predator's and agent's location, and count this event as a ``caught" point for the agent. After the agent gets caught, we continue the simulation. In this case, the agent is required to escape from the predator. Otherwise, it can be caught again. As in the previous scenario, we perform this process for five independent trials, with each trial starting with a randomly initialized ANN.

In this case, we aim to find plasticity rules that train the agents to maximize the number of ``collected'' points and minimize the number of ``caught" points. We combine these two objectives into one equation, as follows:
\begin{equation}
fitness =\frac{1}{T} \sum_{k=1}^{T}(\alpha i_{k} - c_{k})
\label{eq:fitnessPreyPredator}
\end{equation}
\noindent{}where $T=5$ is the number of trials, and $c_{k}$ and $i_{k}$ are the number of ``caught'' and ``collected'' points in the trial $k$, respectively. We use $\alpha$ to increase the weight of the ``collected'' point because $c_{k}$ tend to be larger than $i_{k}$. %The final fitness is the average of a number of trials.

We use a similar network structure to that described in Section~\ref{sec:foragingEnvironment}. However, since this task may require a larger vision to keep track of the movement of preys and predators, we increase the field of vision of the agent to include all the cells in a $9 \times 5$ rectangle ($4$ cells distance on the left, right and in front) in front of the agent. Therefore, the agent is able to see $44$ cells in total, excluding its own location. Since we encode each cell using two bits, we have $88$ inputs to the network. We use one hidden layer with $50$ neurons and $3$ output neurons.

In this case, we define reinforcement signals based on the behavior of the agent with respect to the closest object (either a wall, a prey or a predator). If the closest object in the visual range of the agent is a prey and the agent avoids it by choosing an action that increases its distance to it, then we provide a punishment. Otherwise, if the agent performs an action that reduces its distance to the prey, then we provide a reward. Similar signals are associated to predators and walls. The complete behavior-reinforcement signal associations can be found in Appendix~\ref{appx:sub:reinforcementFunction:preyPredator}.

Similarly to the foraging task, we employ two seasons where we switch the roles of the prey and predator agents. During the summer season, preys are shown in green and predators are shown in blue; during the winter, predators are shown in blue and preys are shown in green. We switch the reinforcement signals in each season accordingly. In total, we run two seasons (summer, winter), each consisting of $4000$ action steps.
\section{Experimental Results}
\label{sec:results}
We present now the results of the ESP rules on the two tasks described above, and compare them with the Hill Climbing (HC) algorithm~\citep{de2006fundamentals}. The HC is an offline optimization approach where a single individual, encoding the connections of an ANN, is evaluated on the task without any lifetime learning capability, and optimized iteratively for a certain number of iterations (see Appendix~\ref{appx:HC} for further details). 

%%%%%%%%%%%%%%%%%%%%%%%%%%%%%%%%%%%%%%%%%%%%%%%%%%%%%%%%%%%%%%%%%%%%

\subsection{Foraging Task}
\label{sec:results_foraging}
In Table~\ref{tab:comparisonForaging}, a comparison of the average fitness results of the best agents controlled/trained by various algorithms is given. In the table, ``Perfect agent'' (PA) refers to the results of an agent controlled by hand-coded rules (i.e., not controlled by an ANN). In this case, the agent has ``perfect knowledge'' about the problem from the beginning of the trial (therefore, there is no lifetime learning).
\vspace{-0.4cm}
\begin{table}[!ht]
\caption{Foraging task: average fitness results of the agents controlled/trained by different algorithms. The details of rules ID:1 and ID18 can be found in Table~\ref{tab:distinctEvolvedRules} and \ref{tab:expertDrivenRules}.} \label{tab:comparisonForaging}
\small
\centering
\begin{tabular}{|l|l|l|l|}
\hline
\textbf{Algorithm}                         & \textbf{Fitness} & \textbf{Std} & \textbf{Learning Type}     \\ \hline
Perfect Agent (PA)                              & 67               & 8.28               & Hand-coded \\ \hline
Hill Climbing (HC)                             & 59               & 19.70              & Offline optimization \\ \hline
Evolved Synaptic Plasticity (rule ID:1)           & 50               & 9.91               & Lifetime learning    \\ \hline
Discrete Hebbian/anti-Hebbian (rule ID:18) & 0.2              & 6.13               & Lifetime learning    \\ \hline
\end{tabular}
\end{table}

We collected a total of $300$ ESP rules by taking the top $10$ best performing rules from $30$ independent GA runs. The overall results of the best performing ESP rule is indicated as ``rule ID:1'' in Table~\ref{tab:comparisonForaging}. The performance of the ESP rule is lower relative to the PA and HC. This is due to lifetime learning. Even though the agents are tested for the same number of action steps, the agent with the ESP rule is required to learn the task during its lifetime. Moreover, the mistakes it makes during this process are included into the fitness evaluation.

The complete collection of ESP rules found by the GA results in $15$ \emph{distinct} rules (out of $3^8$ possible rules), i.e. rules that differ only for their discrete parts (as shown in Figure~\ref{fig:framework}, the discrete parts of the rules consist in $8$-dimensional ternary strings, without considering the specific value of the learning rate). The results of these $15$ distinct rules are given in Table~\ref{tab:distinctEvolvedRules}, where the rules are ranked based on their average fitness values (lower ID rule means better). In the second column of the table, we show how many rule instances we found for each of these rule types. Slightly more than half of the ESP rules (164 out of 300) converged to the first rule type (rule ID:1). The remaining columns, labelled as ``Median'', ``Std'', ``Max'', ``Min'', ``$\eta$ Mean'', and ``$\eta$ Std'', show the median, standard deviation, max and min values of the fitness, and the average and standard deviation of the learning rates for each distinct rule, respectively. The discrete parts of the rules are shown in the remaining eight columns of each row. Values $\{-1,0,1\}$ in each cell indicate what kind of change is performed (\textit{\{decrease, stable, increase\}}, respectively). The column labels indicate the activation states of pre- and post-synaptic neurons ($a_i$ and $a_j$), encoded as 2-bits, and $m=-1$ and $m=1$ indicate the reinforcement signals. For instance, the best performing rule (ID:1) performs the following synaptic changes: when $m=-1$ and $a_i = 1, a_j = 0$, \textit{increase} ($1$); when $m=-1$ and $a_i = 1, a_j = 1$ \textit{decrease} ($-1$); otherwise, the weight is kept \textit{stable} ($0$).

For comparison, Table~\ref{tab:expertDrivenRules} provides the results of three plasticity rules defined by hand. In particular, the rule ID:16 was defined by taking the best performing ESP rule (ID:1) and replacing the case $m=-1$ and $a_i = 1, a_j = 0$ with \textit{stable} ($0$). After this change, the rule performs synaptic changes only when pre- and post-synaptic neurons are active and the network produces an undesired outcome ($m=-1$). As for the other two rules, the rule ID:18 (also shown in the last row in Table~\ref{tab:comparisonForaging}) performs Hebbian/anti-Hebbian learning, as it increases/decreases the synaptic weights between neurons when they are both active and the network produces a desired/undesired outcome. Instead, the rule ID:17 performs synaptic changes on two additional activation combinations w.r.t. rule ID:18, in order to facilitate the creation of new connections. 

From Table~\ref{tab:expertDrivenRules}, we can see that the performance obtained by the rule ID:16 is significantly better than the other two rules (ID:17 and ID:18). However, this performance is worse than the original ESP rule ID:1. Surprisingly though, the GA did not find the rule ID:16, as shown in the distinct rule list given in Table~\ref{tab:distinctEvolvedRules}, even though it performs better than most of all the other rules. This may be due to the convergence of the evolutionary process to the rules that perform better, specifically the rules ID:1, ID:2 and ID:3. As for the rules ID:17 and ID:18, we observe that even though the rule ID:17 performs better than the rule ID:18, its performance is not better than the worst ESP rule (ID:15).

We also observe that the ESP rules that perform synaptic changes when $m=1$ are worse than the ones that perform synaptic changes only when $m=-1$ (ESP rules ID:1, which represents more than half of total $300$ rules, ID:2 and ID:3). This may be due to the design of the reward function for this task. The reward function provides rewards based on the desirable behavior of the network. However, if the plasticity rule continues to perform synaptic changes when the network has already learnt the task, then these changes can disrupt the weights, causing the network to ``forget''. The issue of forgetting already learnt knowledge/skills in ANNs, due to the acquisition of new knowledge/skills, is usually referred to as ``catastrophic forgetting"~\citep{parisi2019continual}.

\begin{landscape}

\begin{table}[ht]
\begin{center}
\footnotesize
\caption{Foraging task: results of the distinct ESP rules found by the GA, ranked by their median fitness. The columns ``Rules'' shows the number of ESP rules found for each distinct rule; ``Median'', ``Std'', ``Max'', ``Min'' show their median, standard deviation, max and minimum fitness, respectively; ``$\eta$ Mean'', ``$\eta$ Std'' show their average learning rate and standard deviations, respectively. The next four columns indicate the plasticity rules \textit{\{decrease, stable, increase\}}, as $\{-1,0,1\}$), in the case of $m=-1$. The last four columns indicate the plasticity rules in the case of $m=1$. The 2-bit headers of the last $8$ columns indicate the binary states of the pre- and post-synaptic neuron activations $a_i$ and $a_j$.}
\label{tab:distinctEvolvedRules}
\begin{tabular}{llllllll|l|l|l|l|l|l|l|l|}
%\cline{9-16} & & & & & & & &

\cline{2-16} & \multicolumn{7}{|c|}{\textbf{Fitness}} & \multicolumn{4}{c|}{$\boldsymbol{m = -1}$} & \multicolumn{4}{c|}{$\boldsymbol{m = 1}$}\\ \hline

\multicolumn{1}{|l|}{\textbf{ID}} &\multicolumn{1}{l|}{\textbf{Rules}} & \multicolumn{1}{l|}{\textbf{Median}} & \multicolumn{1}{l|}{\textbf{Std}} & \multicolumn{1}{l|}{\textbf{Max}} & \multicolumn{1}{l|}{\textbf{Min}} & \multicolumn{1}{l|}{$\boldsymbol{\eta}$\textbf{ Mean}} & $\boldsymbol{\eta}$\textbf{Std} & \textbf{00} & \textbf{01} & \textbf{10} & \textbf{11} & \textbf{00} & \textbf{01} & \textbf{10} & \textbf{11} \\ \hline

\multicolumn{1}{|l|}{\textbf{1}}& \multicolumn{1}{c|}{164}& \multicolumn{1}{l|}{48.63}&\multicolumn{1}{l|}{0.83}&\multicolumn{1}{l|}{49.96}& \multicolumn{1}{l|}{42.95}& \multicolumn{1}{l|}{0.0375}&0.008 &0&0&1&-1&0&0&0&0             \\ \hline

\rowcolor[gray]{0.95}

\multicolumn{1}{|l|}{\textbf{2}}&\multicolumn{1}{c|}{23}& \multicolumn{1}{l|}{44.23}&\multicolumn{1}{l|}{1.05}&\multicolumn{1}{l|}{45.88}& \multicolumn{1}{l|}{41.55}& \multicolumn{1}{l|}{0.0167}&0.004 & -1&1&1&-1&0&0&0&0            \\ \hline

\multicolumn{1}{|l|}{\textbf{3}}&\multicolumn{1}{c|}{3}& \multicolumn{1}{l|}{42.35}&\multicolumn{1}{l|}{2.65}&\multicolumn{1}{l|}{46.45}& \multicolumn{1}{l|}{41.48}& \multicolumn{1}{l|}{0.0192}&0.003 & 1&0&1&-1&0&0&0&0           \\ \hline

\rowcolor[gray]{0.95}
\multicolumn{1}{|l|}{\textbf{4}}&\multicolumn{1}{c|}{19}& \multicolumn{1}{l|}{28.35}&\multicolumn{1}{l|}{0.51}&\multicolumn{1}{l|}{29.05}& \multicolumn{1}{l|}{27.22}& \multicolumn{1}{l|}{0.0488}&0.009 & -1&1&1&-1&1&0&-1&0           \\ \hline

\multicolumn{1}{|l|}{\textbf{5}}&\multicolumn{1}{c|}{1}& \multicolumn{1}{l|}{27.28}&\multicolumn{1}{l|}{0}&\multicolumn{1}{l|}{27.28}& \multicolumn{1}{l|}{27.28}& \multicolumn{1}{l|}{0.0182}&0 & -1&-1&1&0&1&0&-1&0          \\ \hline

\rowcolor[gray]{0.95}
\multicolumn{1}{|l|}{\textbf{6}}&\multicolumn{1}{c|}{9}& \multicolumn{1}{l|}{26.70}&\multicolumn{1}{l|}{1.48}&\multicolumn{1}{l|}{27.91}& \multicolumn{1}{l|}{22.80}& \multicolumn{1}{l|}{0.0118} &0.003& -1&1&1&-1&-1&-1&1&1 \\ \hline

\multicolumn{1}{|l|}{\textbf{7}}&\multicolumn{1}{c|}{16}& \multicolumn{1}{l|}{26.54}&\multicolumn{1}{l|}{0.89}&\multicolumn{1}{l|}{28.13}& \multicolumn{1}{l|}{24.65}& \multicolumn{1}{l|}{0.0092}&0.002& 0&1&1&-1&-1&-1&1&1 \\ \hline

\rowcolor[gray]{0.95}
\multicolumn{1}{|l|}{\textbf{8}}&\multicolumn{1}{c|}{2}& \multicolumn{1}{l|}{25.91}&\multicolumn{1}{l|}{0.07}&\multicolumn{1}{l|}{25.97}& \multicolumn{1}{l|}{25.86}& \multicolumn{1}{l|}{0.0096} &0.0008 &1&1&1&-1&-1&-1&1&1 \\ \hline

\multicolumn{1}{|l|}{\textbf{9}}&\multicolumn{1}{c|}{1}& \multicolumn{1}{l|}{23.55}&\multicolumn{1}{l|}{0}&\multicolumn{1}{l|}{23.55}& \multicolumn{1}{l|}{23.55}& \multicolumn{1}{l|}{0.0273} &0& -1&1&1&-1&0&-1&0&1 \\ \hline

\rowcolor[gray]{0.95} 
\multicolumn{1}{|l|}{\textbf{10}}&\multicolumn{1}{c|}{2}& \multicolumn{1}{l|}{22.11}&\multicolumn{1}{l|}{1.26}&\multicolumn{1}{l|}{23}& \multicolumn{1}{l|}{21.21}& \multicolumn{1}{l|}{0.0052}&0.003& 1&1&1&-1&0&-1&0&1 \\ \hline
    
\multicolumn{1}{|l|}{\textbf{11}}&\multicolumn{1}{c|}{10}& \multicolumn{1}{l|}{20.96}&\multicolumn{1}{l|}{0.33}&\multicolumn{1}{l|}{21.59}& \multicolumn{1}{l|}{20.45}& \multicolumn{1}{l|}{0.0198}&0.003& 0&1&1&-1&1&-1&-1&-1 \\ \hline
   
\rowcolor[gray]{0.95}  
\multicolumn{1}{|l|}{\textbf{12}}&\multicolumn{1}{c|}{20}& \multicolumn{1}{l|}{20.63}&\multicolumn{1}{l|}{0.40}&\multicolumn{1}{l|}{21.34}& \multicolumn{1}{l|}{19.81}& \multicolumn{1}{l|}{0.061} &0.022& -1&1&1&0&1&-1&-1&-1 \\ \hline
   
\multicolumn{1}{|l|}{\textbf{13}}&\multicolumn{1}{c|}{1}& \multicolumn{1}{l|}{12.41}&\multicolumn{1}{l|}{0}&\multicolumn{1}{l|}{12.41}& \multicolumn{1}{l|}{12.41} & \multicolumn{1}{l|}{0.0799}&0& 0&0&0&-1&1&1&0&-1 \\ \hline

\rowcolor[gray]{0.95}  
\multicolumn{1}{|l|}{\textbf{14}}&\multicolumn{1}{c|}{19}& \multicolumn{1}{l|}{10.30}&\multicolumn{1}{l|}{0.55}&\multicolumn{1}{l|}{10.87}& \multicolumn{1}{l|}{8.36}& \multicolumn{1}{l|}{0.0662} &0.018& 0&0&0&-1&0&1&1&-1 \\ \hline
   
\multicolumn{1}{|l|}{\textbf{15}}&\multicolumn{1}{c|}{10}& \multicolumn{1}{l|}{8.82}&\multicolumn{1}{l|}{0.54}&\multicolumn{1}{l|}{9.59}& \multicolumn{1}{l|}{7.97}& \multicolumn{1}{l|}{0.0301}&0.009& 1&1&-1&-1&1&0&-1&1 \\ \hline
\end{tabular}
\end{center}
%\end{table}

%\end{landscape}

%\begin{landscape}
%\begin{table}[ht]
\begin{center}
\footnotesize
\caption{Foraging task: results (fitness values, standard deviations of the fitness values, number of correctly and incorrectly collected items and their standard deviations, and number of wall hits and its standard deviation) of three rules defined by hand, over $100$ trials. Columns encoded as 2-bits represent the activation states of pre- and post-synaptic neurons.} \label{tab:expertDrivenRules}
\begin{tabular}{llllllllll|l|l|l|l|l|l|l|l|}
\cline{11-18} & & & & & & & & & \textbf{}  & \multicolumn{4}{c|}{$\boldsymbol{m = -1}$} & \multicolumn{4}{c|}{$\boldsymbol{m = 1}$} \\ \hline

\multicolumn{1}{|l|}{\textbf{ID}} & \multicolumn{1}{l|}{\textbf{Fitness}} & \multicolumn{1}{l|}{\textbf{Std Fit.}} & \multicolumn{1}{l|}{\textbf{Correct}} & \multicolumn{1}{l|}{\textbf{Std Cor.}} & \multicolumn{1}{l|}{\textbf{Incor.}} & \multicolumn{1}{l|}{\textbf{Std Inc.}} & \multicolumn{1}{l|}{\textbf{Wall}} & \multicolumn{1}{l|}{\textbf{Std Wall}} & $\boldsymbol{\eta}$ & \textbf{00} & \textbf{01} & \textbf{10} & \textbf{11} & \textbf{00} & \textbf{01} & \textbf{10} & \textbf{11} \\ \hline

\multicolumn{1}{|l|}{\textbf{16}} & \multicolumn{1}{l|}{41.68}	& \multicolumn{1}{l|}{18.26}	& \multicolumn{1}{l|}{47.63}	& \multicolumn{1}{l|}{14.36}	& \multicolumn{1}{l|}{5.95}	& \multicolumn{1}{l|}{7.17}	& \multicolumn{1}{l|}{14.36}	& \multicolumn{1}{l|}{54.59}	& \multicolumn{1}{l|}{0.04}	&0&	0&	0&	-1&	0&	0&	0&	0 \\ \hline

\rowcolor[gray]{0.95}
\multicolumn{1}{|l|}{\textbf{17}} & \multicolumn{1}{l|}{5.6}	& \multicolumn{1}{l|}{9.28}	& \multicolumn{1}{l|}{25.2}	& \multicolumn{1}{l|}{6.88}	& \multicolumn{1}{l|}{19.7}	& \multicolumn{1}{l|}{5.28}	& \multicolumn{1}{l|}{120}	& \multicolumn{1}{l|}{140.2}	& \multicolumn{1}{l|}{0.01}	&0&	0&	1&	-1&	0&	-1&	0&	1 \\ \hline

\multicolumn{1}{|l|}{\textbf{18}} & \multicolumn{1}{l|}{0.2}	& \multicolumn{1}{l|}{6.13}	& \multicolumn{1}{l|}{18.3}	& \multicolumn{1}{l|}{4.48}	& \multicolumn{1}{l|}{18.1}	& \multicolumn{1}{l|}{4.27}	& \multicolumn{1}{l|}{602.6}	& \multicolumn{1}{l|}{117.5}	& \multicolumn{1}{l|}{0.01}	&0&	0&	0&	-1&	0&	0&	0&	1 \\ \hline

\end{tabular}
\end{center}
\end{table}

\end{landscape}

\begin{figure*}[ht]
\begin{subfigures}
\subfloat[Rule ID:1]{\includegraphics[width=0.52\columnwidth]{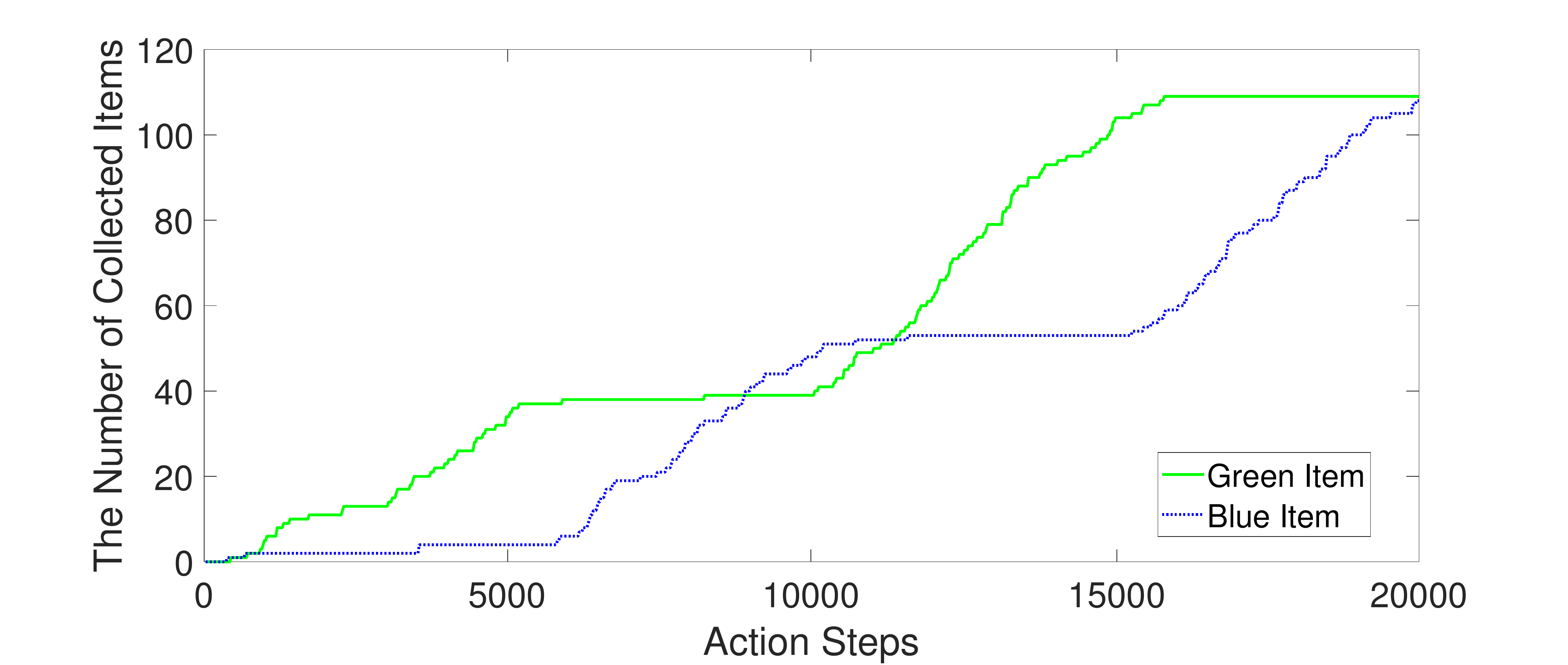}
\label{fig:CLseasonDuring1}}
\subfloat[Rule ID:6]{\includegraphics[width=0.52\columnwidth]{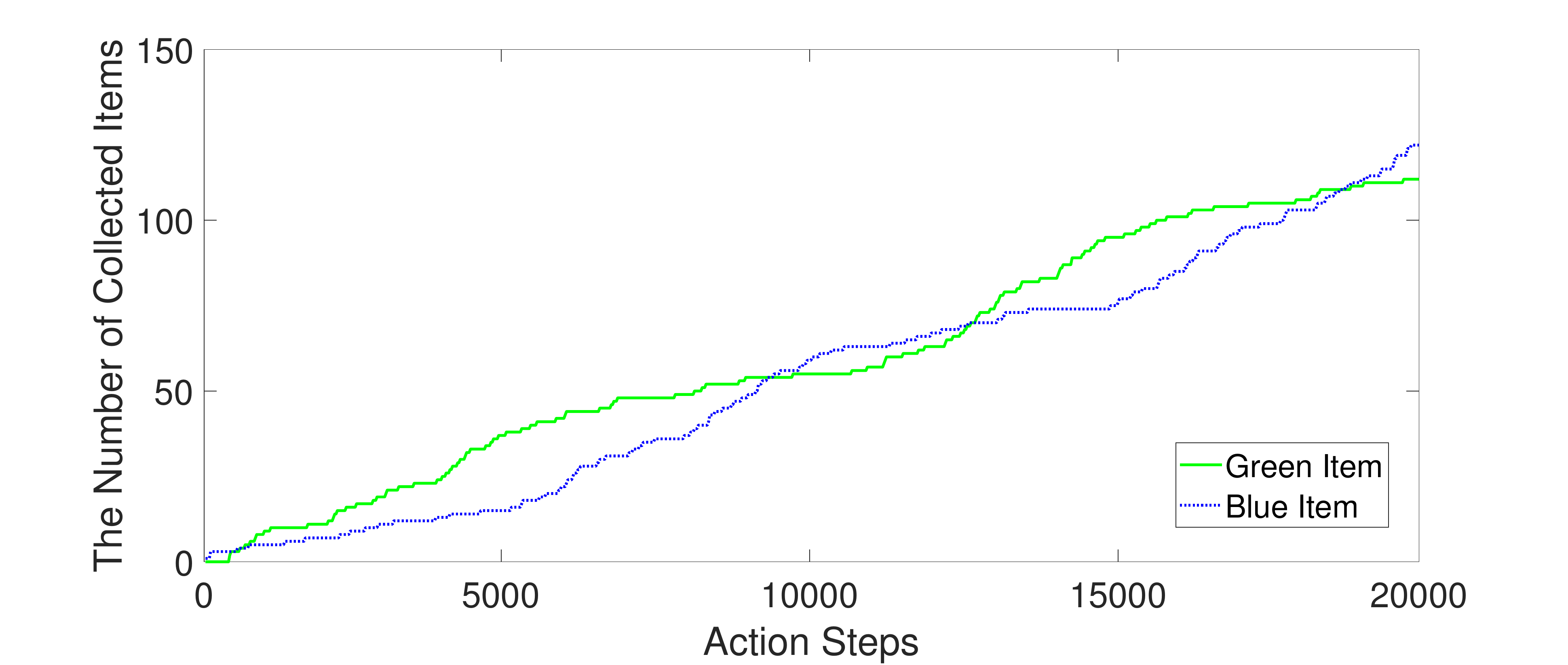}
\label{fig:CLseasonDuring2}}
\end{subfigures}
\caption{Foraging task: number of green and blue items collected with the two ESP rules ID:1 (a) and ID:6 (b), see Table~\ref{tab:distinctEvolvedRules} for details.}%The season is changed every $5000$ steps.
\label{fig:continuousResultsDisplay}
\end{figure*}

Finally, in Figure~\ref{fig:continuousResultsDisplay} we report the number of collected items during a lifetime learning process of agents with the ESP rules ID:1 and ID:6. Every $5000$ steps, the season is changed. We observe that the agent with rule ID:6 is not as efficient as the agent with rule ID:1 at adapting to the seasonal changes. In fact, the agent with rule ID:1 learns to avoid the incorrect item faster than the agent with rule ID:6. We present additional results of the agents during the foraging task in Appendix~\ref{appx:sub:PerformanceDuring}\footnote{To gain further insight into these results, we have visually inspected the behavior of the agent with ESP rule ID:1 and $\eta=0.0375$ during a foraging task with four seasons, each consisting of $3000$ action steps (we reduced the season duration for visualization purposes). The video (available at \url{https://youtu.be/9jy6yTFKgT4}) shows that the agent is capable of efficiently adapting to the environmental conditions imposed by each season. Even though the desired behavior with respect to the wall should be constant across different seasons, the agent makes a few mistakes at the beginning of each season by hitting the wall. This may be due to the change of the synaptic weights that affect the behavior of the agent with respect to the wall. We provide the weights and their change after each seasonal change in Appendix~\ref{appx:sub:ChangeOfSynapticWeights}.}.

%%%%%%%%%%%%%%%%%%%%%%%%%%%%%%%%%%%%%%%%%%%%%%%%%%%%%%%%%%%%%%%%%%%%

\subsection{Prey-predator Task}
\label{sec:prey_predator}
The average results of the best agents controlled/trained using HC and best ESP rule on the prey-predator task are shown in Table~\ref{tab:preyPredatorComparisonResults}. In this case we collected $70$ best performing ESP rules, running $7$ independent runs of the GA. We show these rules in Table~\ref{tab:distinctEvolvedRulesPreyPredator} (summarized based on their discrete parts, as we did for the foraging task in Table~\ref{tab:distinctEvolvedRules}).

\begin{table}[ht]
\caption{Prey-predator task: average fitness results of the agents controlled/trained by Hill Climbing and the best ESP rule (ID:1 in Table~\ref{tab:distinctEvolvedRulesPreyPredator}).}
\label{tab:preyPredatorComparisonResults}
\small
\centering
\begin{tabular}{|l|l|l|l|}
\hline
\textbf{Algorithm} & \textbf{Fitness} & \textbf{Std} & \textbf{Learning Type} \\ \hline
Hill Climbing (HC) & 41.9 & 16.43 & Offline optimization \\ \hline
Evolved Synaptic Plasticity (rule ID:1) & 31.88 & 22.88 & Lifetime learning \\ \hline
\end{tabular}
\end{table}

\begin{table}[ht]
\begin{center}
%\footnotesize
%\scriptsize
\caption{Prey-predator task: results of the distinct ESP rules found by the GA, ranked by their median fitness. Columns are labelled as in Table~\ref{tab:distinctEvolvedRules}.}
\label{tab:distinctEvolvedRulesPreyPredator}
\resizebox{\textwidth}{!}{
\begin{tabular}{llllllll|l|l|l|l|l|l|l|l|}
\cline{2-16} & \multicolumn{7}{|c|}{\textbf{Fitness}} & \multicolumn{4}{c|}{$\boldsymbol{m = -1}$}                  & \multicolumn{4}{c|}{$\boldsymbol{m = 1}$}                   \\ \hline
\multicolumn{1}{|l|}{\textbf{ID}} &\multicolumn{1}{l|}{\textbf{Rules}} & \multicolumn{1}{l|}{\textbf{Median}} & \multicolumn{1}{l|}{\textbf{Std}} & \multicolumn{1}{l|}{\textbf{Max}} & \multicolumn{1}{l|}{\textbf{Min}} & \multicolumn{1}{l|}{$\boldsymbol{\eta}$\textbf{ Mean}} & $\boldsymbol{\eta}$\textbf{Std} & \textbf{00} & \textbf{01} & \textbf{10} & \textbf{11} & \textbf{00} & \textbf{01} & \textbf{10} & \textbf{11} \\ \hline

\multicolumn{1}{|l|}{\textbf{1}}&\multicolumn{1}{c|}{10}& \multicolumn{1}{l|}{24.20}&\multicolumn{1}{l|}{4.39}&\multicolumn{1}{l|}{29.46}& \multicolumn{1}{l|}{16.12}& \multicolumn{1}{l|}{0.42} &0.16& 0&-1&1&0&1&-1&0&0 \\ \hline

\rowcolor[gray]{0.95}
\multicolumn{1}{|l|}{\textbf{2}}& \multicolumn{1}{c|}{35}& \multicolumn{1}{l|}{23.84}&\multicolumn{1}{l|}{3.06}&\multicolumn{1}{l|}{31.88}& \multicolumn{1}{l|}{19.21}& \multicolumn{1}{l|}{0.55}&0.06 &0&-1&1&0&-1&0&-1&-1             \\ \hline

\multicolumn{1}{|l|}{\textbf{3}}&\multicolumn{1}{c|}{1}& \multicolumn{1}{l|}{21.57}&\multicolumn{1}{l|}{0}&\multicolumn{1}{l|}{21.57}& \multicolumn{1}{l|}{21.57}& \multicolumn{1}{l|}{0.98}&0 & 1&-1&1&-1&0&-1&-1&-1            \\ \hline

\rowcolor[gray]{0.95}
\multicolumn{1}{|l|}{\textbf{4}}&\multicolumn{1}{c|}{14}& \multicolumn{1}{l|}{16.54}&\multicolumn{1}{l|}{2.28}&\multicolumn{1}{l|}{20.42}& \multicolumn{1}{l|}{13.42}& \multicolumn{1}{l|}{0.69}&0.16 &  1&0&1&-1&-1&0&-1&-1     \\ \hline

\multicolumn{1}{|l|}{\textbf{5}}&\multicolumn{1}{c|}{7}& \multicolumn{1}{l|}{10.65}&\multicolumn{1}{l|}{2.98}&\multicolumn{1}{l|}{15.26}& \multicolumn{1}{l|}{6.56}& \multicolumn{1}{l|}{0.56}&0.08 & 
1&0&1&-1&1&-1&-1&0          \\ \hline

\rowcolor[gray]{0.95}
\multicolumn{1}{|l|}{\textbf{6}}&\multicolumn{1}{c|}{3}& \multicolumn{1}{l|}{10.61}&\multicolumn{1}{l|}{1.93}&\multicolumn{1}{l|}{11.93}& \multicolumn{1}{l|}{8.39}& \multicolumn{1}{l|}{0.65}&0.02 & 1&0&1&-1&0&-1&-1&0          \\ \hline
\end{tabular}
}
\end{center}
\end{table}

Similarly to the foraging task, the results of the ESP rule are worse than those of HC, due to lifetime learning. On the other hand, we note that in this case the ESP rules converge to larger learning rates (one order of magnitude higher compared to the foraging task). This is probably due to the increased stochasticity of this task: since preys and predators move randomly, it is difficult for the agent to learn their behavior patterns. Therefore, the agent needs to adjust its behavior very quickly.

Also in this case, we have visually inspected the behavior of the agent. In particular, we focused on the two best ESP rules (ID:1 and ID:2 in Table~\ref{tab:distinctEvolvedRulesPreyPredator}, respectively with $\eta=0.3550$ and $\eta=0.4377$)\footnote{A video recording of the behaviors of the agents controlled/trained by HC and the two ESP rules is accessible online at \url{https://youtu.be/97vWjTrTEzE}.}. Quite interestingly, we observed that the behaviors obtained by the two best ESP rules are quite different, even though they have similar results in terms of median fitness. The behavior obtained with rule ID:1 is in fact similar to that obtained with HC: the agent moves straight until it encounters an object. On the other hand, the plasticity rule ID:2 appears to have evolved in a way to perform synaptic changes that modifies the behavior of the agent to move diagonally. This behavior appears to be advantageous against the movements of the preys and predators.
\section{Conclusions}
\label{sec:conclusion}
The plasticity property of biological and artificial neural networks enables learning by modifying the networks' configurations. These modifications take place at individual synapse/neuron level, based on local interactions between neurons.

In this work, we proposed an evolutionary approach to optimize/discover synaptic plasticity rules to produce autonomous learning in changing environmental conditions. We represented the plasticity rules in a binary form, to perform changes based on pairwise binary activations of neurons. Most of the works in literature consider evolving complex functions to perform synaptic changes. However, the binary representation used in this work can reduce the search space of possible rules and allow for interpretation. This may provide insights into the learning behavior of the networks for certain learning scenarios. Accordingly, we presented the ESP rules discovered in this work and discussed their behaviors.

We evaluated the proposed algorithm on agent-based foraging and prey-predator tasks. We focused specifically on the adaptation capabilities of the ANNs in the cases where the environmental conditions change. To demonstrate this, we defined two seasons that are associated with different reinforcement signals, and measured the lifetime adaptation capability of the networks during these seasonal changes.

We collected the best performing ESP rules after running the GAs multiple times. These rules converged into several types and differed between foraging and prey-predator tasks. For instance, in the case of the foraging task, the best ESP rule performed synaptic changes only when the network produced undesired output (negative reinforcement signal). This is likely due to the reward functions we used in the experimentation, which were designed to provide constant reward/punishment while the networks produced desired/undesired outcomes. Intuitively, after the networks learn to perform the task successfully, continuing to perform synaptic changes may cause degradation in the synaptic weights and result in forgetting. 

In the case of prey-predator task, the best ESP rules were more complex than they were in the foraging task. In contrast, the best ESP rules tend to perform frequent synaptic changes to adapt to the stochasticity of the prey-predator task. 

To set an upper bound for the performance of the ANNs with ESP rules, we performed a set of separate experiments using hand-coded rule-based agents and ANN controllers optimized using the HC algorithm. Comparison with these algorithms in foraging and prey-predator tasks showed that the agents trained with ESP rules could perform the task very well (as good as about $74\%$ of the performance of the HC), considering the continuous learning versus offline optimization.

In future work, we aim to investigate the scalability of the ESP rules to larger networks and various learning tasks. Furthermore, as encountered in this work, we are also particularly interested in investigating the approaches to avoid catastrophic forgetting in continuous learning scenarios.

\section*{Acknowledgements}
\noindent
\begin{tabular}{p{0.10\linewidth} p{0.8\linewidth}}
\raisebox{-0.8cm}{\includegraphics[height=.95cm]{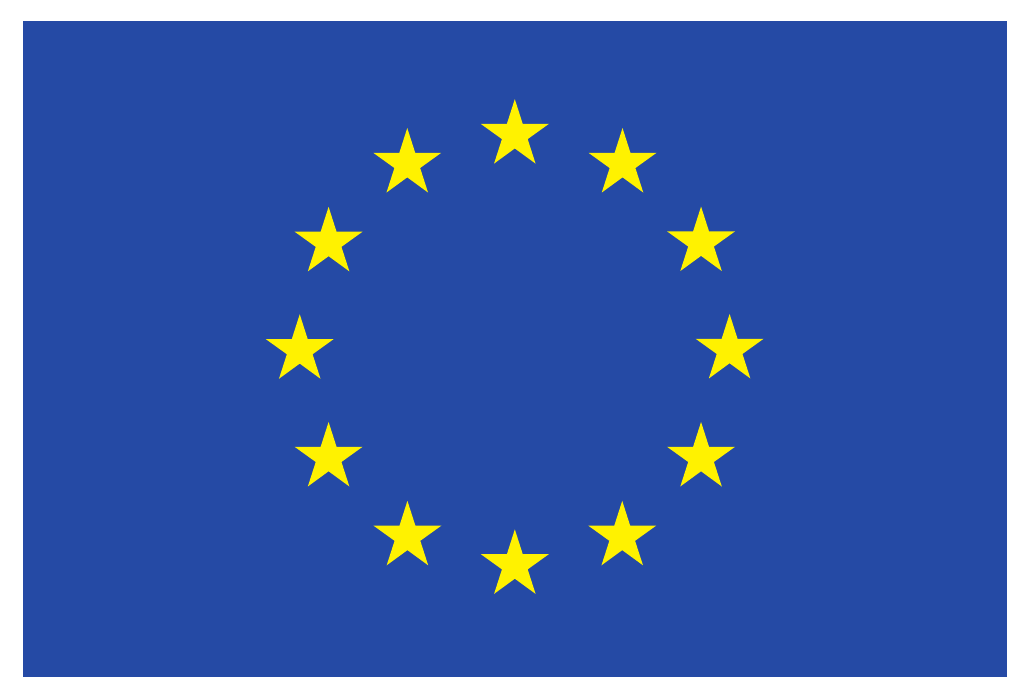}}
&
{This project has received funding from the European Union's Horizon 2020 research and innovation programme under grant agreement No: 665347.}
\end{tabular}

%\newpage
\bibliographystyle{apalike}

\newpage
\appendix
\section{Artificial Neural Network Model}
\label{appx:anns}
ANNs consist of a number of \textit{artificial neurons} arranged within a certain connectivity pattern, where a directed connection between two neurons is referred to as a \textit{synapse}, a neuron with an outgoing synapse is referred to as a \textit{pre-synaptic neuron}, and a neuron with an incoming synapse is referred to as a \textit{post-synaptic neuron}.

In our experiments, we use fully connected feed-forward ANN (see Figure~\ref{fig:agentANN}), in which the activation of each post-synaptic neuron, $a_i$, is calculated as:
\begin{equation}
a_i = \psi 
\left( 
\sum_{j=0} w_{ij} a_j 
\right) 
\label{eq:activationPostSynaptic}
\end{equation}
where $a_j$ is the activation of the $j$-th pre-synaptic neuron, $w_{i,j}$ is the synaptic efficiency (weight) from pre-synaptic neuron $j$ to post-synaptic neuron $i$, $a_0$ is a \emph{bias}, which usually takes a constant value of $1$, and $\psi()$ is the activation function, that in our case is set to a \emph{step activation function} given by:
\begin{equation}
\psi(x) = \left\{
 \begin{array}{ll}
  1, & \mbox{if }x>0\mbox{;}\\
  0, & \mbox{otherwise.}
 \end{array}
\right. 
\label{eq:activationFunction}
\end{equation}
The step function reduces the possible activation states of each neuron into two possibilities: \textit{active} (1) and \textit{passive} (0). As discussed below, this binary neuron model facilitates the interpretability of the results.

%%%%%%%%%%%%%%%%%%%%%%%%%%%%%%%%%%%%%%%%%%%%%%%%%%%%%%%%%%%%%%%%%%%%

\section{Hill Climbing Algorithm}
\label{appx:HC}
In the case of the HC algorithm, we employ an ANN with the same architecture used in the ESP experiments, whose weights are randomly initialized in $[-1,1]$. We evaluate the ANN on the first season of the task, to find the initial best network and its corresponding fitness value. We then iteratively generate a candidate network by randomly perturbing all the weights of the best network using a Gaussian mutation with $0$ mean $0.1$ standard deviation. The newly created network is evaluated on the next season of the task, and replaces the best network if its fitness value is better. We repeat this process iteratively for all the seasons of the task (in our experiments, $4$ alternating summer/winter seasons for the foraging task, and $2$ summer/winter seasons for the prey-predator task). In both tasks, we consider in this case $1000$ action steps per season. When a seasonal change happens, we keep the best network and continue the optimization procedure as specified. The final fitness result is the average of $100$ independent HC optimization process.

%%%%%%%%%%%%%%%%%%%%%%%%%%%%%%%%%%%%%%%%%%%%%%%%%%%%%%%%%%%%%%%%%%%%

\section{Behavior-Reinforcement Signal Associations}
\label{appx:reinforcementFunction}

\subsection{Foraging Task}
\label{appx:sub:reinforcementFunction:foraging}
The complete list of sensory states, behaviors and reinforcement signal associations that we used in the foraging task is provided in Table~\ref{tab:rewardFunction}. In the table, the columns labelled as ``Sensor'' and ``Behavior'' show the sensor states and actions of the network, respectively. The reward functions corresponding to the two seasons, ``Summer'' and ``Winter'', are also shown. In the following, green items, blue items and walls are referred to as ``G'', ``B'' and ``W'' respectively.
%These are defined with the expectation to observe the emergence of certain behaviors.

It should be noted that for some sensory states, multiple reward function associations may be triggered. In these cases, the associations that concern the behaviors to collect/avoid items are given priority. For instance, when the sensory input of the agent indicates that there are ``W straight" and ``G on left", and if the agent goes right, we activate the reward association ID:12, rather than ID:3. We should also note that the reward functions described here do not specify the reinforcement signal outcomes for \emph{all} possible sensor-behavior combinations. Indeed, in total there are $192$ possible sensor-behavior associations (resulting from 3 possible behavior outcomes for each of $2^6$ possible sensor states). We assume that all the sensor-behavior associations that are not shown in the table do no provide any reinforcement signal.

The first two reward associations are defined to encourage the agent to explore the environment. Otherwise, the agent may get stuck in a small area, for example by performing only actions such as going left or right when there is nothing present. The sensor state labelled as ``nothing'' refers to an input to the network equal to $[0, 0, 0, 0, 0, 0]$.

The reward associations from ID:3 to ID:8 specify the reinforcement signals for the behaviors with respect to the wall, and are the same in both summer and winter seasons. It is expected that the agent learns to avoid the wall. Therefore, we define positive reward signals for the states where there is a wall, and the agent picks a behavior that avoids a collision (IDs: 3, 5, 7). Conversely, we define punishment signals for the sensor-behavior associations where the agent collides to the wall (IDs: 4, 6, 8). For instance, the sensor state of the association given in ID:3 refers to the input to the network as $[0, 0, 1, 1, 0, 0]$, and provides reward if the agent decides to go left or right (and, this behavior is desired in both summer and winter seasons).

The reward associations from ID:9 to ID:14, and from ID:15 to ID:20, define the reinforcement signals associated to the ``G'' and ``B'' items respectively. The reinforcement signals are reversed between the two seasons. In summer, the agent is expected to collect ``G'' items and avoid ``B'' items, whereas in winter the agent is expected to collect ``B'' items and avoid ``G'' items.

\begin{table}[!ht]
\begin{center}
\small
\caption{Foraging task: associations of the sensor and behavior states to the reinforcement signals in the two seasons.}
\vspace{0.1cm}
\label{tab:rewardFunction}
\begin{tabular}{|l|l|l|c|c|}
\hline
\textbf{ID}    & \textbf{Sensor} & \textbf{Behavior} & \textbf{Summer} & \textbf{Winter} \\ \hline
\textbf{1}   & nothing  & Straight  & 1& 1 \\ \hline
\textbf{2}   & nothing  & Left or Right  & -1& -1 \\ \hline \hline
\textbf{3}   & W straight  & Left or Right  & 1& 1 \\ \hline
\rowcolor[gray]{0.95}
\textbf{4}   &  W straight  & Straight  & -1 & -1 \\ \hline
\textbf{5}   &  W on left & Right  & 1& 1 \\ \hline
\rowcolor[gray]{0.95}
\textbf{6}   &  W on left & Left or Straight  & -1 & -1 \\ \hline
\textbf{7}   &  W on right & Left  & 1& 1 \\ \hline
\rowcolor[gray]{0.95}
\textbf{8}   &  W on right & Right or Straight  & -1& -1 \\ \hline \hline
\textbf{9}   &  G straight & Straight  & 1 & -1 \\ \hline
\rowcolor[gray]{0.95}
\textbf{10}   & G straight & Left or Right  & -1 & 0 \\ \hline
\textbf{11}   & G on left & Left  & 1 & -1 \\ \hline
\rowcolor[gray]{0.95}
\textbf{12}   & G on left & Straight or Right  & -1 & 0 \\ \hline
\textbf{13}   & G on right & Right  & 1& -1 \\ \hline
\rowcolor[gray]{0.95}
\textbf{14}   & G on right & Straight or Left  & -1 & 0 \\ \hline \hline
\textbf{15}   & B straight & Straight  & -1 & 1 \\ \hline
\rowcolor[gray]{0.95}
\textbf{16}   & B straight & Left or Right  & 0 & -1 \\ \hline
\textbf{17}   & B on left & Left  & -1 & 1 \\ \hline
\rowcolor[gray]{0.95}
\textbf{18}   & B on left & Straight or Right  & 0 & -1 \\ \hline
\textbf{19}   & B on right & Right  & -1 & 1 \\ \hline
\rowcolor[gray]{0.95}
\textbf{20}   & B on right & Straight or Left  & 0 & -1 \\ \hline
\end{tabular}    
\end{center}
\end{table}

%%%%%%%%%%%%%%%%%%%%%%%%%%%%%%%%%%%%%%%%%%%%%%%%%%%%%%%%%%%%%%%%%%%%

\subsection{Prey-predator Task}
\label{appx:sub:reinforcementFunction:preyPredator}
The complete list of sensory states, behaviors and reinforcement signal associations that we used in the prey-predator task is provided in Table~\ref{tab:rewardFunctionPreyPredator}. The header labelled as ``Closest'' refers to the object that is the closest to the agent (in its visual range), measured using Euclidean distance. The behavior labelled as ``Avoid'' refers to the situation where the agent selects an action that maximizes its distance to the closest object. On the contrary, the behavior labelled as ``Move towards'' refers to the situation where the agent selects an action that minimizes its distance to the closest object. 

The first four reward associations are similar to those used in the foraging task. In particular, the first two associations (ID:1 and ID:2) encourage exploration, while the reward associations related to the wall (ID:3 and ID:4) also encourage (punish) behaviors that avoid (move towards) the wall. 

As for the behavior with respect to the preys and predators, reflected in the reward associations from ID:5 to ID:8, during the summer season the agent is required to avoid blue agents (predators) and collect green agents (preys). During the winter season, the prey-predator roles are switched. Therefore, the agent is required to avoid green agents (predators), and collect blue agents (predators).
%The header ``Behavior'' refers to the action of the agent. The agent can only perform one of ``Left'', ``Straight'', or ``Right'' actions. 
%
\vspace{-0.5cm}
\begin{table}[!ht]
\begin{center}
\small
\caption{Prey-predator task: associations of the sensor and behavior states to the reinforcement signals in the two seasons.}
\label{tab:rewardFunctionPreyPredator}
\begin{tabular}{|l|l|l|c|c|}
\hline
\textbf{ID}    & \textbf{Closest} & \textbf{Behavior} & \textbf{Summer} & \textbf{Winter} \\ \hline
\textbf{1}   & nothing  & Straight  & 1& 1\\ \hline
\textbf{2}   & nothing  & Right or Left & -1& -1 \\ \hline \hline
\textbf{3}   & W & Avoid  & 1& 1 \\ \hline
\rowcolor[gray]{0.95}
\textbf{4}   &  W & Move towards  & -1 & -1 \\ \hline \hline
\textbf{5}   &  G & Avoid & -1 & 1\\ \hline %(prey) 
\rowcolor[gray]{0.95}
\textbf{6}   &  G & Move towards & 1 & -1 \\ \hline \hline %(prey) 
\textbf{7}   &  B & Avoid & 1  & -1 \\ \hline %(predator) 
\rowcolor[gray]{0.95}
\textbf{8}   &  B & Move towards  & -1 & 1 \\ \hline %(predator) 
\end{tabular}    
\end{center}
\end{table}

%%%%%%%%%%%%%%%%%%%%%%%%%%%%%%%%%%%%%%%%%%%%%%%%%%%%%%%%%%%%%%%%%%%%

\section{Additional Results on the Foraging Task}
\label{appx:ForagingTask}

\subsection{Detailed Results of the Hill Climbing Algorithm}
\label{appx:sub:HCAdditionalResults}
\begin{figure}[!ht]
\begin{center}
\includegraphics[width=\columnwidth]{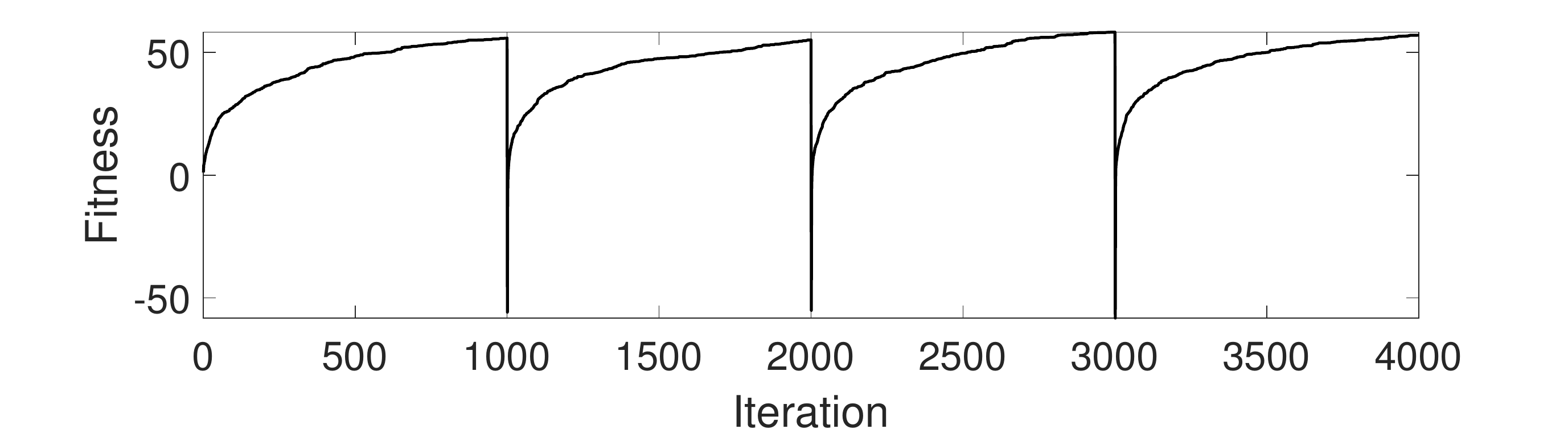}
\caption{Foraging task: average fitness value of $100$ runs of the offline optimization process of the agents using the HC algorithm. The process starts with a randomly initialized ANN, starting from the summer season. Every $1000$ iterations, the season is changed, for a total of $4$ seasons (see Table~\ref{tab:HCResults}).}
\label{fig:HC4seasons}
\end{center}
\vspace{-0.2cm}
\end{figure}

Figure~\ref{fig:HC4seasons} shows the average fitness value of $100$ runs of the optimization process with the HC algorithm. Every $1000$ iterations, the season is switched. In the first $1000$ iterations, HC is able to find an agent that performs the task efficiently, with a fitness value of around $55$. When the season is switched to winter (1001st iteration), the agent still performs according to the summer season, thus achieving a fitness score of $-55$ (since the expected behavior is reversed). After about $1000$ iterations, HC is able to find an agent that performs the task in the new season efficiently. We observe a similar behavior in the following seasonal changes.  

In Table~\ref{tab:HCResults}, we report the mean and standard deviations of the fitness values, the collected number of green and blue items, and the wall hits at the 10th, 500th and 1000th iteration for each season. The values for Summer1, Winter1, Summer2 and Winter2 presented in the table correspond, respectively, to the iteration ranges 0-1000, 1001-2000, 2001-3000 and 3001-4000 on the $x$-axis of Figure~\ref{fig:HC4seasons}.
\begin{table}[!ht]
\caption{Foraging task: average and standard deviations of the results of 100 runs of the offline optimization procedure of the agents using the Hill Climbing algorithm.}
\label{tab:HCResults}
\vspace{-0.3cm}
%\small
\begin{center}
\resizebox{\textwidth}{!}{
\begin{tabular}{|l|l|l|l|l|l|}
\hline
\textbf{Result} & \textbf{Iteration} & \textbf{Summer1} & \textbf{Winter1} & \textbf{Summer2} & \textbf{Winter2} \\ \hline  
\multirow{3}{*}{\textbf{Fitness}} & \textbf{10th} & $9.20\pm   10.12$   & $9.79  \pm 11.83$   & $8.74 \pm  10.24$   &$10.41\pm   12.04$\\ \cline{2-6} 
                                  & \textbf{500th} & $48.34 \pm  24.82$  & $47.42  \pm 19.15$   &$49.59 \pm  18.02$ &  $49.94  \pm 17.04$ \\ \cline{2-6} 
                                  & \textbf{1000th} & $55.81  \pm 23.04$ &  $55.08  \pm 20.00$  & $58.21 \pm  17.08$&   $56.95 \pm  18.35$\\ \hline
\multirow{3}{*}{\textbf{G}}       & \textbf{10th} &  $11.61  \pm 12.37$   & $5.02\pm    7.63$  & $13.32  \pm 12.22$  & $4.48\pm    6.46$ \\ \cline{2-6} 
                                  & \textbf{500th} &$49.36  \pm 24.48$   & $1.58  \pm  4.87$  & $51.71  \pm 17.43$  &  $1.59   \pm 4.62$\\ \cline{2-6} 
                                  & \textbf{1000th}   &$56.56  \pm 22.88$&    $0.85   \pm 3.29$ &  $59.65 \pm  16.89$    &$1.38 \pm   4.09$\\ \hline
\multirow{3}{*}{\textbf{B}}       & \textbf{10th} &$2.41  \pm  5.59$  & $14.81 \pm  14.40$   & $4.58  \pm  6.66$  & $14.89  \pm 14.05$\\ \cline{2-6} 
                                  & \textbf{500th}  &$1.02  \pm 3.23$ &  $49.00  \pm 19.06$&    $2.12  \pm  4.73$  & $51.53  \pm 16.33$\\ \cline{2-6} 
                                  & \textbf{1000th} &$0.75  \pm  1.43$  & $55.93  \pm 19.88$   & $1.44   \pm 3.47$  & $58.33  \pm 17.48$ \\ \hline
\multirow{3}{*}{\textbf{W}}       & \textbf{10th} &$614.09\pm	919.95$&	$693.78\pm	854.64$&	$770.47\pm	1101.77$&	$702.46	\pm 862.67$\\ \cline{2-6}
                                  & \textbf{500th} & $170.26\pm	463.02$&	$157.15\pm	446.59$&	$146.47\pm	417.37$&	$183.05\pm	485.90$\\ \cline{2-6} 
                                  & \textbf{1000th} & $114.92\pm	375.96$&	$88.30\pm 351.09$&	$115.72\pm	398.26$&	$140.97\pm 439.38$ \\ \hline
\end{tabular}
}
\end{center}
\end{table}

We observe that it takes about $1000$ iterations for the networks to reach on average a fitness value above $55$ at the end of each season. Since the networks are well optimized for the task, a sudden decrease in their performance at the beginning of each seasonal change is observed. Still, after $10$ iterations in all seasons the agents achieve a fitness value of around $9$. We see clear increasing and decreasing trends in the number of collected items depending on the season, as the number of iterations increases. Moreover, the number of wall hits decreases although at the end of each season it is still relatively high.

%%%%%%%%%%%%%%%%%%%%%%%%%%%%%%%%%%%%%%%%%%%%%%%%%%%%%%%%%%%%%%%%%%%%

\subsection{Performance of the Agents During the Task}
\label{appx:sub:PerformanceDuring}

Figure~\ref{fig:continuousLearningValidationSeasonal} shows the results of some selected ESP rules during a single run of a foraging task over $4$ seasons. The overall process lasts $20000$ action steps in total, consisting of $4$ seasons of $5000$ action steps each. The foraging task starts with the summer season and switches to the other season every $5000$ action steps. Measurements were sub-sampled every $20$ action steps to allow a better visualization. The figures given in the first column show the cumulative number of items of both types collected throughout the process. After each synaptic change, we separately test the network on the same task without continuous learning (i.e., fixing the weights) to show how each synaptic change affects the performance of the network. These results are shown in the figures given in the second column.

\begin{landscape}

\begin{figure}[!ht]
\begin{center}
\begin{minipage}[c]{0.6\textwidth}
\begin{subfigures}
\subfloat[]{\includegraphics[width=0.7\textwidth]{figs/CLseasonalItems1.pdf}\label{fig:CLseasonItems1}}
\subfloat[]{\includegraphics[width=0.7\textwidth]{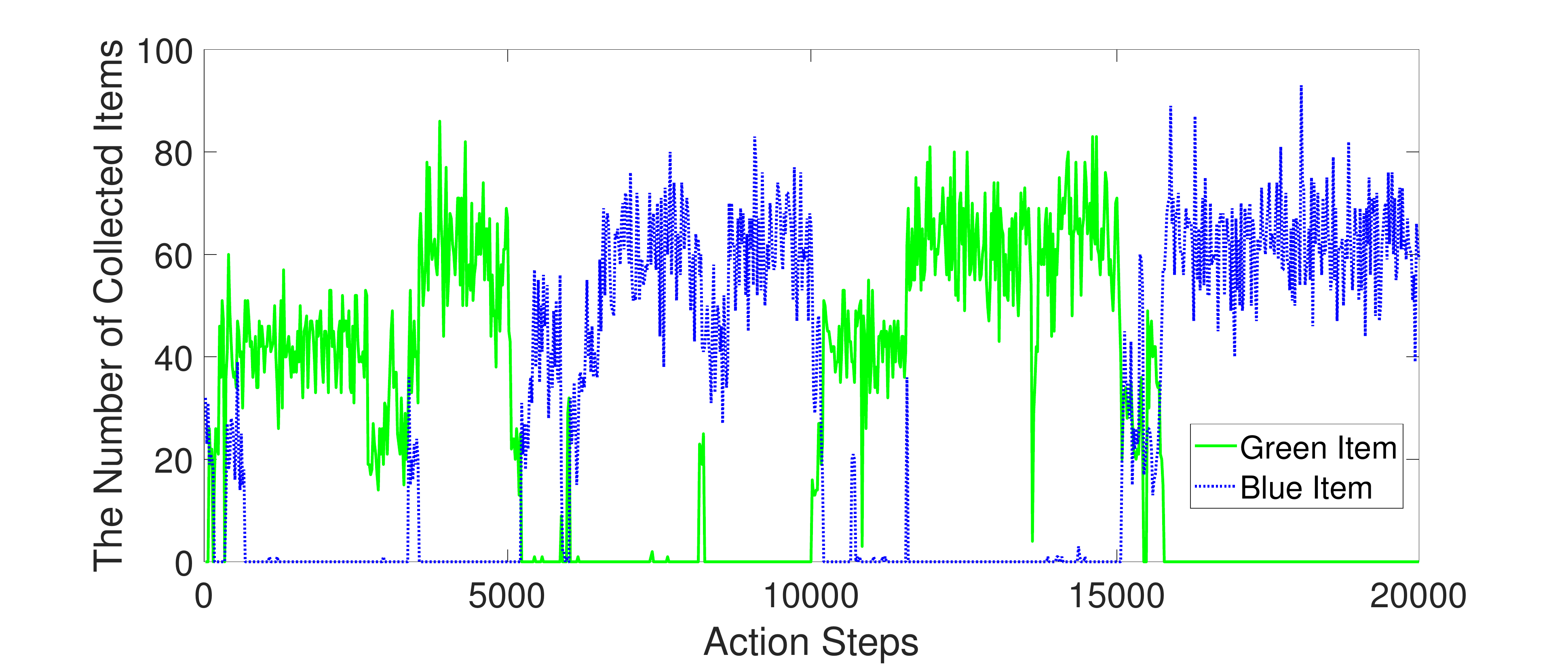}\label{fig:CLseasonal1}}

\subfloat[]{\includegraphics[width=0.7\textwidth]{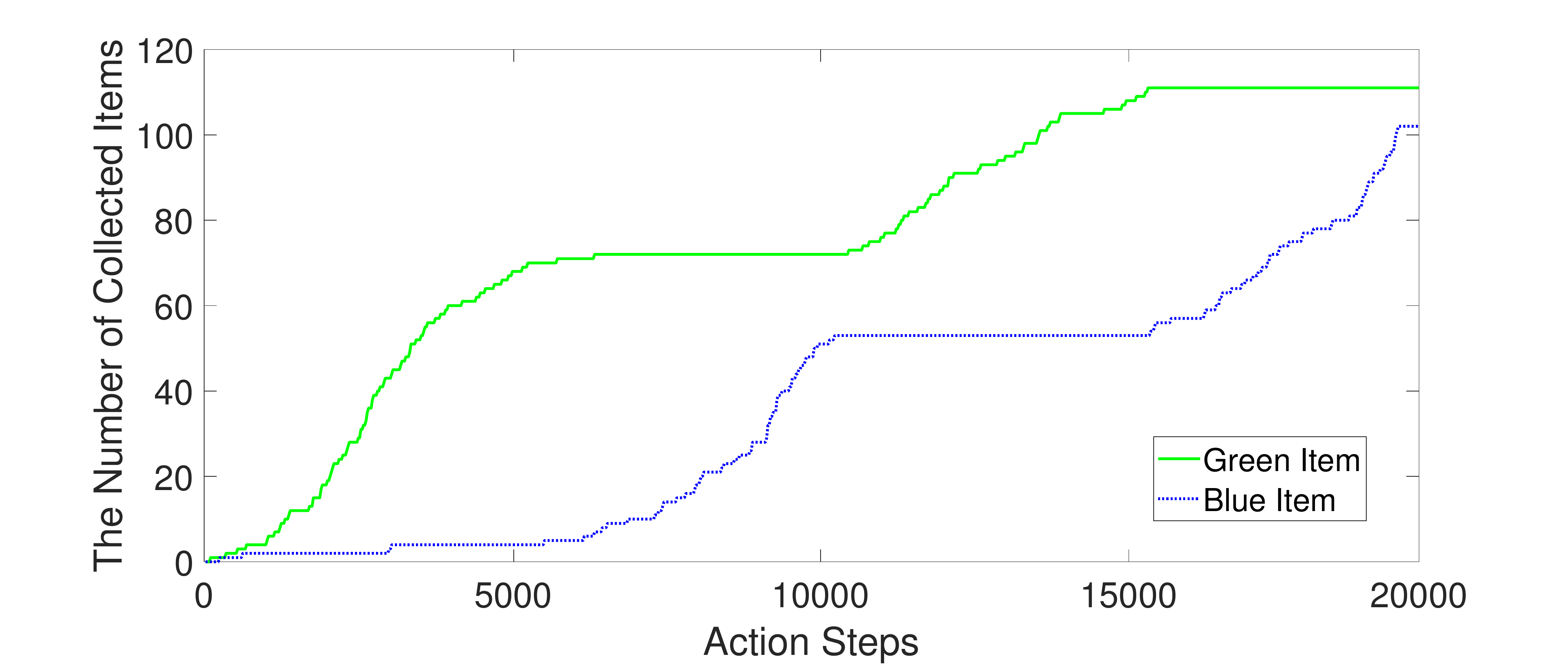}\label{fig:CLseasonItems2}}
\subfloat[]{\includegraphics[width=0.7\textwidth]{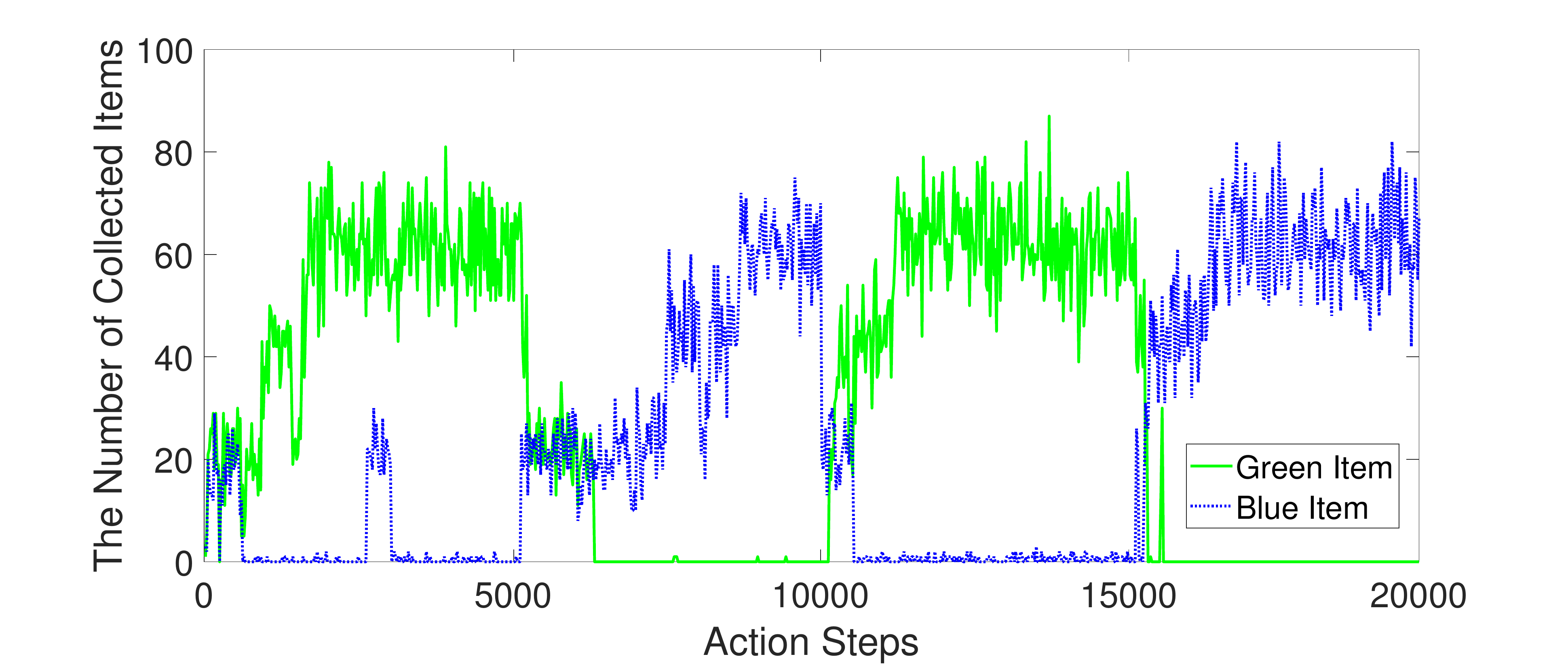}\label{fig:CLseasonal2}}

\subfloat[]{\includegraphics[width=0.7\textwidth]{figs/CLseasonalItems3.pdf}\label{fig:CLseasonItems3}}
\subfloat[]{\includegraphics[width=0.7\textwidth]{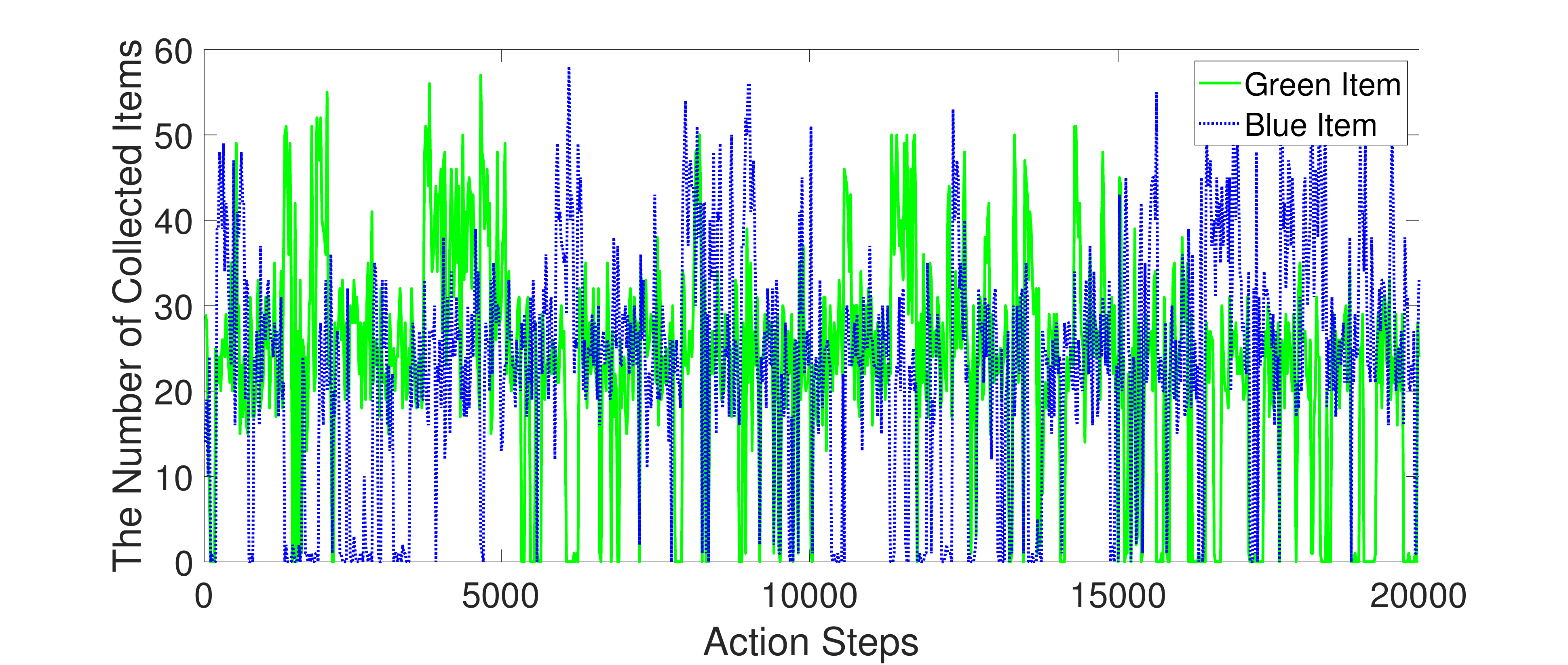}\label{fig:CLseasonal3}}
\end{subfigures}
\end{minipage}
\hspace{3cm}
\begin{minipage}[c]{0.4\textwidth}
\caption{Foraging task: results of three selected ESP rules during a single run of a foraging task over $4$ seasons. Each row of the figures provides the results of the best ESP rules ID:1, ID:2, and ID:6 respectively. Figures in the first column, (a), (c) and (e), show cumulative results on the number of items collected. The configurations of the ANNs are changing during their lifetime. To assess how these changes reflect in behavior, we perform separate tests of these configurations at each action step for $5000$ steps by keeping the weights constant. The number of collected items during these tests are shown in figures (b), (d) and (f) given in the second column.} \label{fig:continuousLearningValidationSeasonal}
\end{minipage}
\end{center}
\end{figure}

\begin{table}[!ht]
\begin{center}
\caption{Foraging task: results of the best ESP rule (ID:1) for each season.}
\label{tab:bestEvolvedRuleStatsContinuous}

\footnotesize
\begin{tabular}{l|l|l|l|l|l|l|l|l|l|l|l|l|l|l|l|l|}
\cline{2-17}
                                         & \multicolumn{4}{c|}{\textbf{Summer1}}                                                                                                      & \multicolumn{4}{c|}{\textbf{Winter1}}                                                                                                      & \multicolumn{4}{c|}{\textbf{Summer2}}                                                                                                      & \multicolumn{4}{c|}{\textbf{Winter2}}                                                                                                      \\ \cline{2-17} 
                                         & \multicolumn{1}{c|}{\textbf{Fitness}} & \multicolumn{1}{c|}{\textbf{G}} & \multicolumn{1}{c|}{\textbf{B}} & \multicolumn{1}{c|}{\textbf{W}} & \multicolumn{1}{c|}{\textbf{Fitness}} & \multicolumn{1}{c|}{\textbf{G}} & \multicolumn{1}{c|}{\textbf{B}} & \multicolumn{1}{c|}{\textbf{W}} & \multicolumn{1}{c|}{\textbf{Fitness}} & \multicolumn{1}{c|}{\textbf{G}} & \multicolumn{1}{c|}{\textbf{B}} & \multicolumn{1}{c|}{\textbf{W}} & \multicolumn{1}{c|}{\textbf{Fitness}} & \multicolumn{1}{c|}{\textbf{G}} & \multicolumn{1}{c|}{\textbf{B}} & \multicolumn{1}{c|}{\textbf{W}} \\ \hline

\multicolumn{1}{|l|}{\textbf{Mean}} & 50.81 &  53.37 & 2.56 & 5.53 & 48 & 4.13 & 52.1 & 3.6 & 50.37 & 53.71 &	3.3& 3.18 & 50.61 & 3.7 & 54.33& 4.7 \\ \hline
         
\multicolumn{1}{|l|}{\textbf{Std}} &10 & 9.5& 1.8&	3.5& 9.2  & 2.2 & 8.4 &	3.6 & 10.2& 9.64&	1.7& 3.4 & 10.2& 1.9 & 9.5 & 6.3 \\ \hline
\end{tabular}
\end{center}
\end{table}

\end{landscape}

\begin{figure}[ht!]
%\begin{minipage}[c]{0.6\textwidth}
\begin{subfigures}
\subfloat[Summer1]{\includegraphics[width=0.48\textwidth]{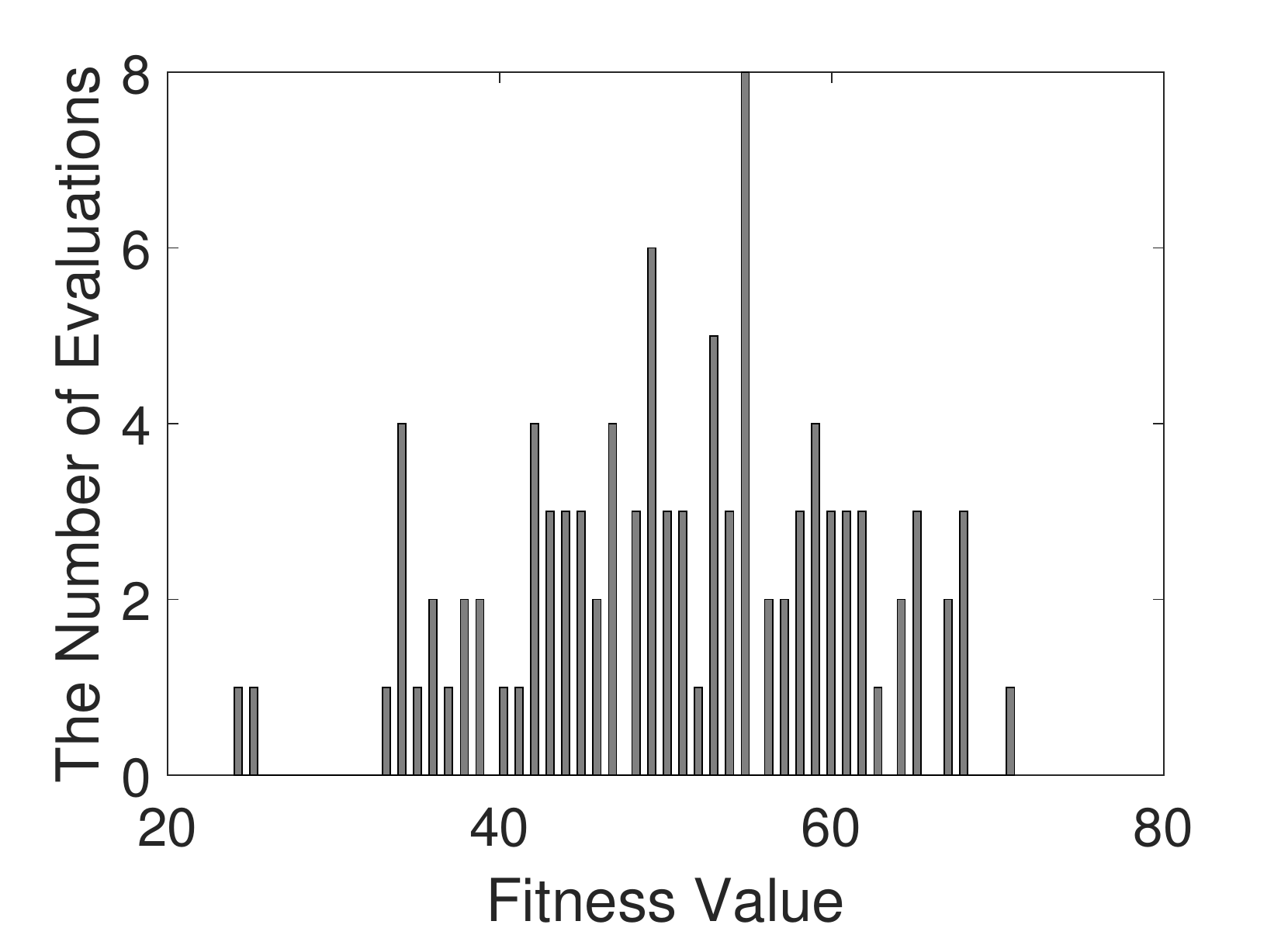}\label{fig:histSummer1}}
\subfloat[Winter1]{\includegraphics[width=0.48\textwidth]{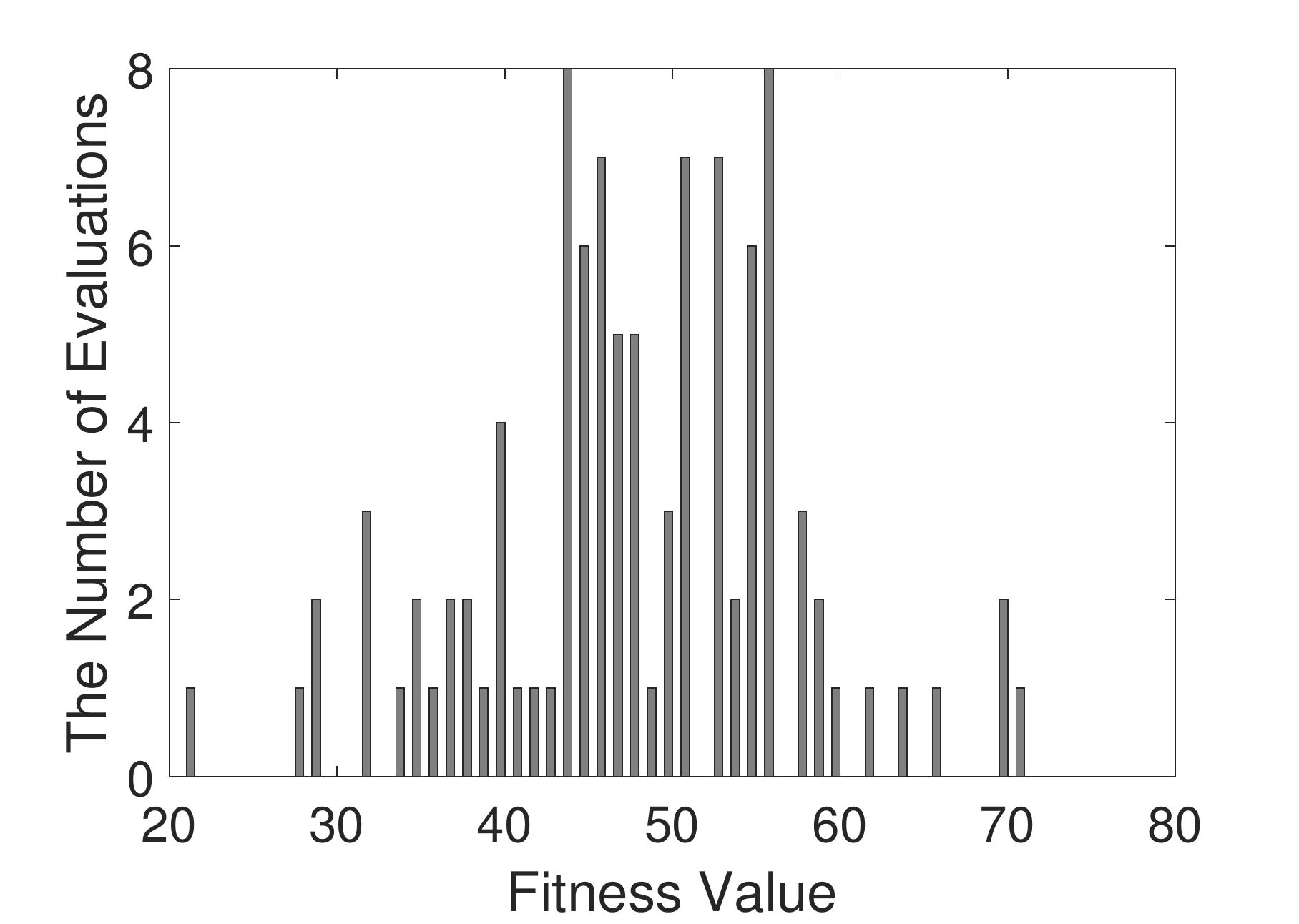}\label{fig:histWinter1}}

\subfloat[Summer2]{\includegraphics[width=0.48\textwidth]{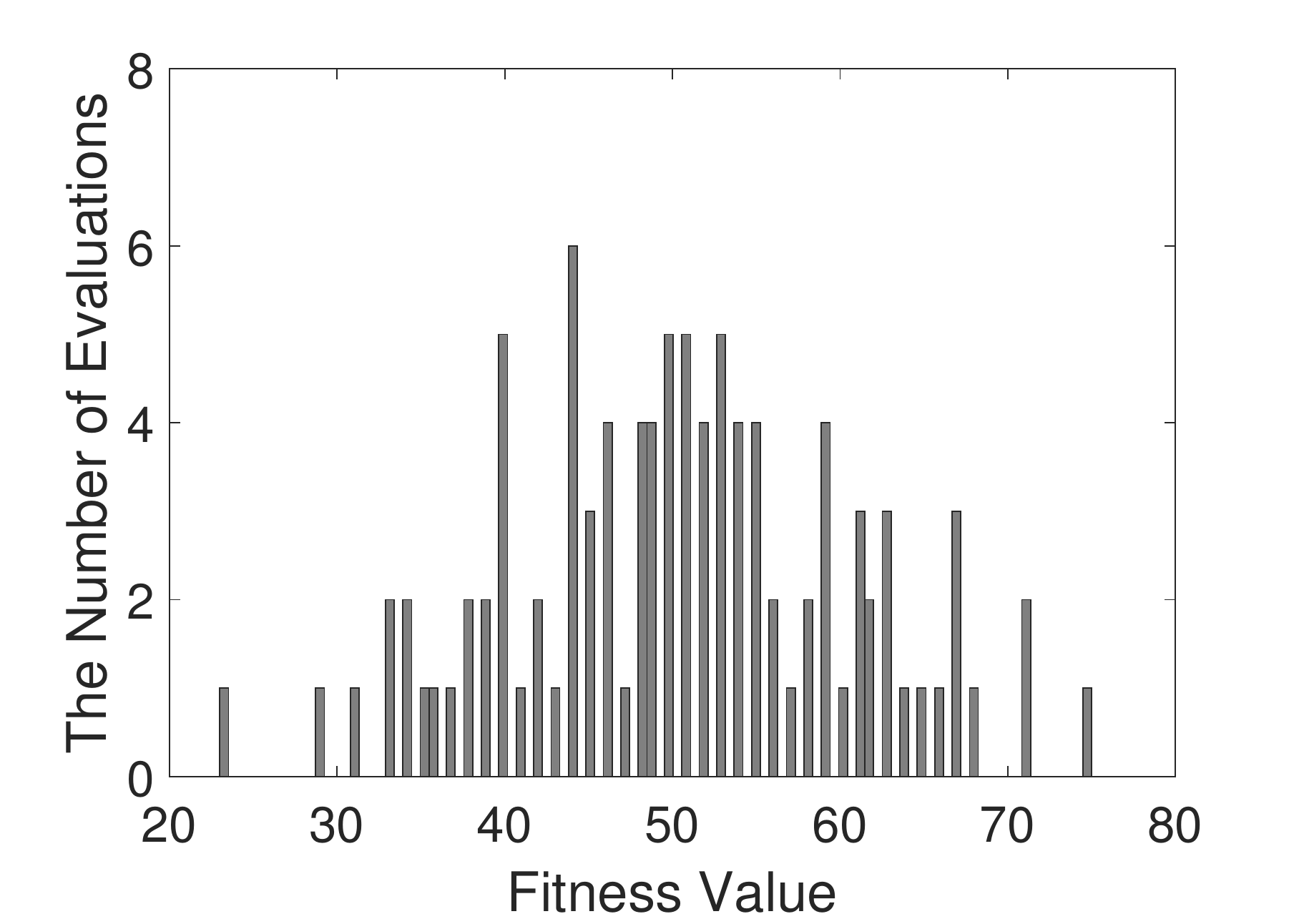}\label{fig:histSummer2}}
\subfloat[Winter2]{\includegraphics[width=0.48\textwidth]{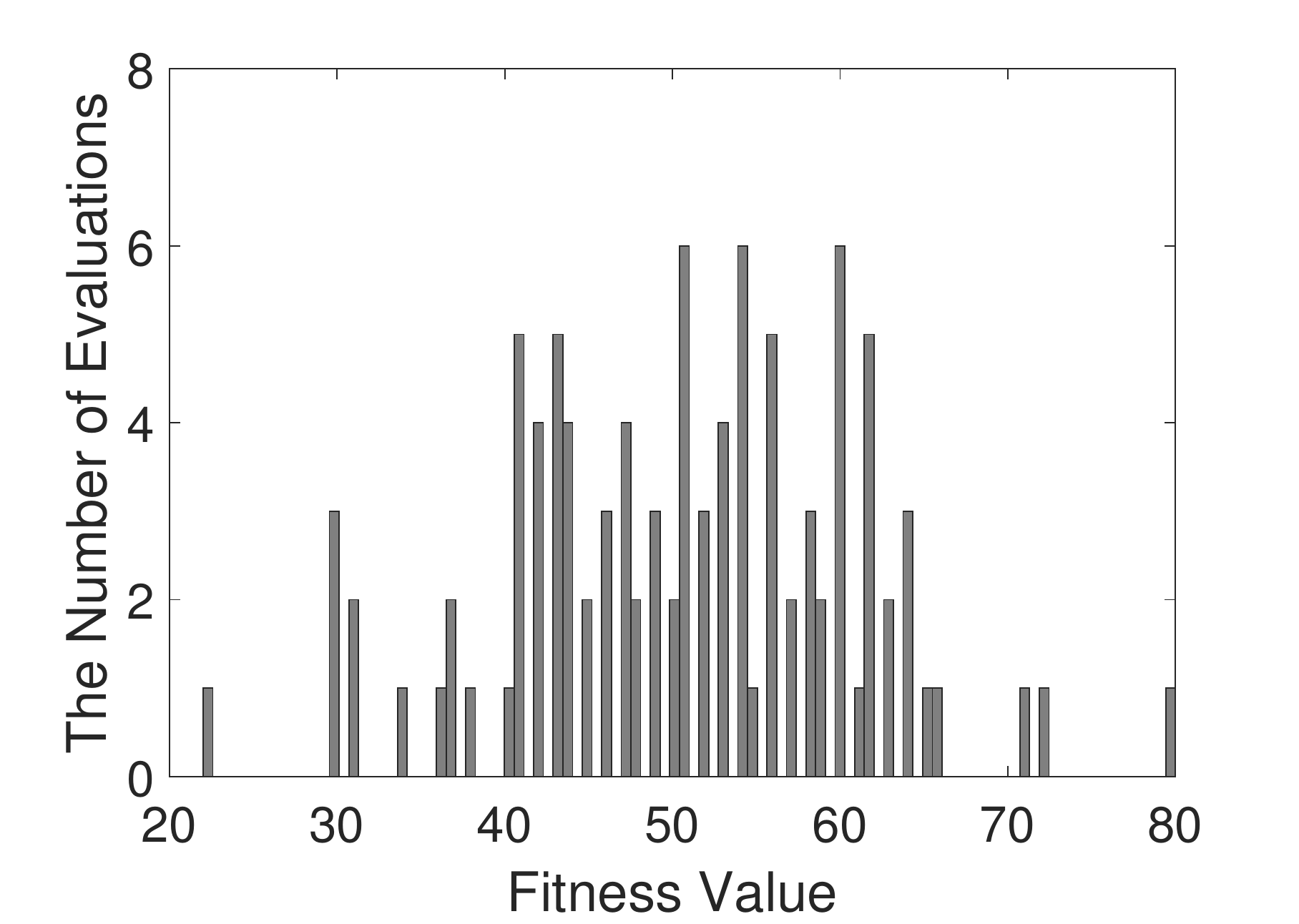}\label{fig:histWinter2}}
\end{subfigures}
%\end{minipage}
%\hfill
%\begin{minipage}[c]{0.25\textwidth}
\caption{Foraging task: distribution of fitness values of the best ESP rule ID:1 over $4$ seasons. The $x$- and $y$-axes of each subfigure show the fitness value and number of evaluations, respectively.} \label{fig:continuousLearningHist}
%\end{minipage}
\end{figure}

More specifically, Figures \ref{fig:CLseasonItems1} and \ref{fig:CLseasonal1} show the results of the best ESP rule ID:1. Figures \ref{fig:CLseasonItems2} and \ref{fig:CLseasonal2} show the results of the best ESP rule ID:2. Figures \ref{fig:CLseasonItems3} and \ref{fig:CLseasonal3} show the results of the best ESP rule ID:6. 

In the case of the ESP rules ID:1 and ID:2, the agent learns quickly to collect the correct type of items in each season. The number of correctly collected items from the previous season stabilizes, and the number of incorrectly collected items from the previous season increases in the next season after a seasonal change occurs. We observe in Figures \ref{fig:CLseasonal1} and \ref{fig:CLseasonal2} that there are distinct and stable seasonal trends for each season. The noise in the measurements is due to the stochasticity of the process.

In the case of the ESP rule ID:6, the agent keeps collecting both kinds of items; however, the number of correctly collected items is larger than the number of incorrectly collected items in each season. We can observe in Figure~\ref{fig:CLseasonal3} that the testing performance is not stable throughout the process. The fluctuations show that the configuration of the ANN seems to be frequently shifting between learning and forgetting.

Table~\ref{tab:bestEvolvedRuleStatsContinuous} and Figure~\ref{fig:continuousLearningHist} show, respectively, the average results and the distribution of the fitness of the best performing ESP rule (ID:1) in each season, over $100$ trials. The agent achieves an average fitness value of about $50$ in all the seasons, except for the first winter season where it achieves a fitness value of $48$. 

\begin{table}[ht]
\caption{Foraging task: results of the best performing ESP rule with validation.}
\label{tab:validation}
\begin{center}
\begin{tabular}{l|l|l|l|l|l|l|l|l|}
\cline{2-9}
                                         & \multicolumn{4}{c|}{\textbf{Summer}}                                                                                                      & \multicolumn{4}{c|}{\textbf{Winter}}                                                                                                       \\ \cline{2-9} 
                                         & \multicolumn{1}{c|}{\textbf{Fitness}} & \multicolumn{1}{c|}{\textbf{G}} & \multicolumn{1}{c|}{\textbf{B}} & \multicolumn{1}{c|}{\textbf{W}} & \multicolumn{1}{c|}{\textbf{Fitness}} & \multicolumn{1}{c|}{\textbf{G}} & \multicolumn{1}{c|}{\textbf{B}} & \multicolumn{1}{c|}{\textbf{W}}  \\ \hline

\multicolumn{1}{|l|}{\textbf{Mean}} & 61.0 &  61.0 & 0 &0 & 63.0 & 0 &63.0 & 0 \\ \hline
        
\multicolumn{1}{|l|}{\textbf{Std}} &8.62& 8.62& 0&	0&  8.19  & 0 &  8.19 &	0 \\ \hline
\end{tabular}
\end{center}
\end{table}

Table~\ref{tab:validation} shows the average results of the best ANN configuration found through validation. In validation, the agent is tested independently every $20$ action steps (the configuration of its ANN is fixed and tested separately on the same task without using continuous learning), and its fitness value at the end of the testing is stored.

Finally, we use the Wilcoxon rank-sum test~\citep{wilcoxon1945} to assess the statistical significance of the difference of the results produced by different algorithms. The null-hypothesis, i.e. that the means of the results produced by two algorithm (thus, their agent's behavior) are the same, is rejected if the $p$-value is smaller than $\alpha = 0.05$. We perform pairwise comparisons of three sets of results of fitness values for summer and winter seasons obtained from three agents evaluated over $100$ evaluations. More specifically, we compare the results of the hand-coded rule-based agent (see the ``Perfect Agent" described in Section~\ref{sec:results_foraging}) with those of the agents that use the best performing ESP rule (ID:1) in continuous learning settings, with and without validation. Based on the pairwise comparison results, the hand-coded rule-based agent is significantly better than the other two agents with and without validation, with significance levels of $1.9\cdot 10^{-28}$ and $8.2\cdot 10^{-05}$ respectively. Furthermore, the results of the agent with validation are significantly better than those without validation, with a significance level of $9.9\cdot 10^{-14}$.

%%%%%%%%%%%%%%%%%%%%%%%%%%%%%%%%%%%%%%%%%%%%%%%%%%%%%%%%%%%%%%%%%%%%

\subsection{Sensitivity Analysis to the Number of Hidden Neurons}
\label{appx:sub:SensitivityNumHiddenNeeurons}
\begin{figure}[!ht]
\begin{center}
\includegraphics[trim=0 7.5cm 0 7.5cm, clip, width = 0.8\textwidth]{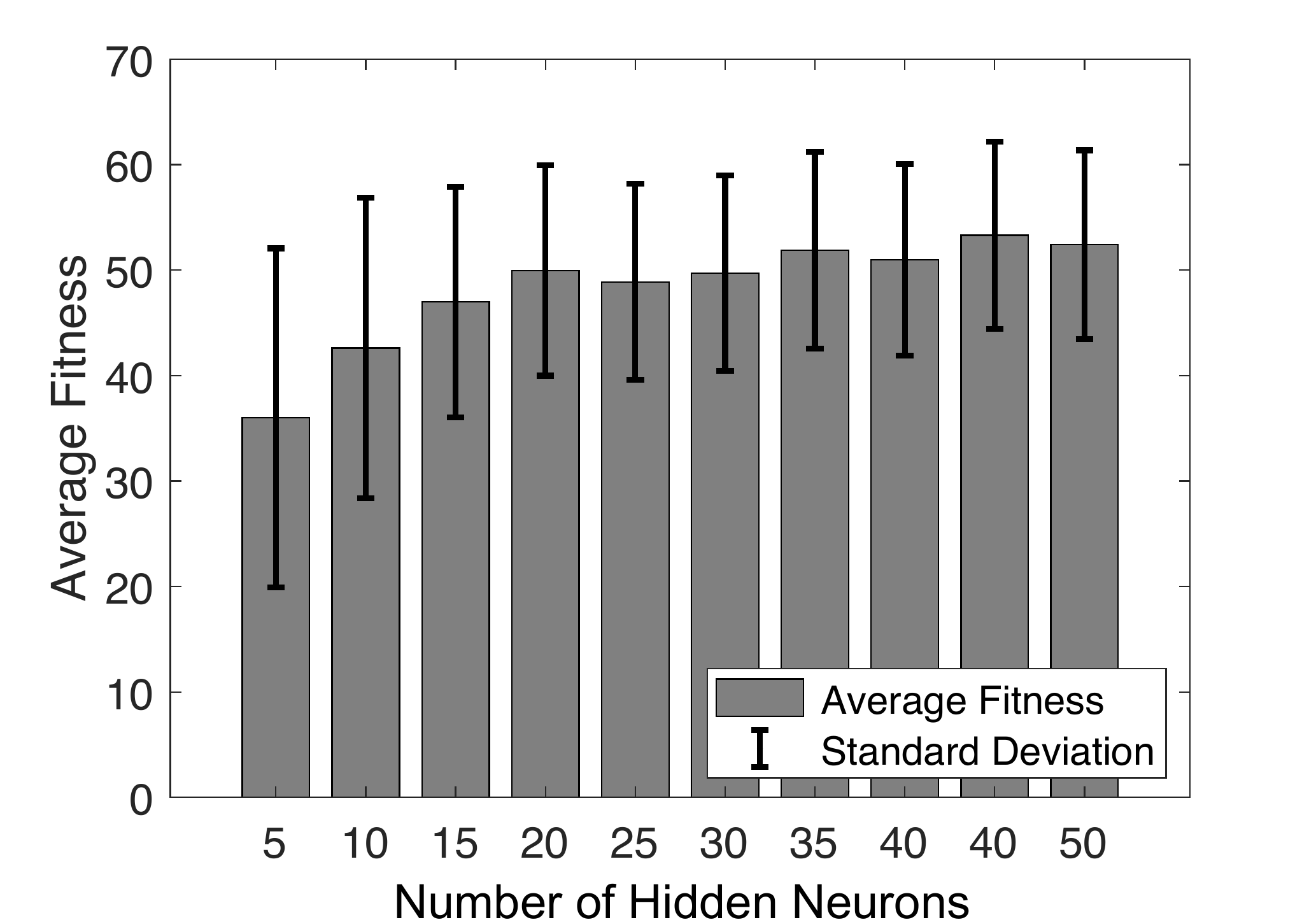}
\caption{Foraging task: average fitness values of the ANNs with various numbers of hidden neurons.}
\label{fig:variousHidden}
\end{center}
\end{figure}

We performed a sensitivity analysis of our results with respect to the number of hidden neurons. In Figure~\ref{fig:variousHidden}, we show the average results of $100$ trials, each performed using the best performing ESP rule (ID:1), with various numbers of hidden neurons. We test networks with $5, 10, 15, 20, 25, 30, 35, 40, 45$ and $50$ hidden neurons. The results show that there is a $13$ point increase on the average fitness value when the number of hidden neurons is increased from $5$ to $20$ (the latter is the valued used in all the experiments reported in the paper). There is also a slight upward trend when more than $20$ hidden neurons are used, which results in a further $3$ point average fitness gain when the number of hidden neurons is increased from $20$ to $50$. Moreover, the standard deviations are also slightly reduced while the number of hidden neurons increases, showing that the use of more hidden neurons tends to make the agent's behavior more consistent across different trials.

%%%%%%%%%%%%%%%%%%%%%%%%%%%%%%%%%%%%%%%%%%%%%%%%%%%%%%%%%%%%%%%%%%%%

\subsection{Change of the Synaptic Weights After Seasonal Changes} \label{appx:sub:ChangeOfSynapticWeights}
\begin{figure*}[ht]
\begin{subfigures}
\subfloat[Randomly sampled initial connection weights.]
{\includegraphics[width=0.5\columnwidth]{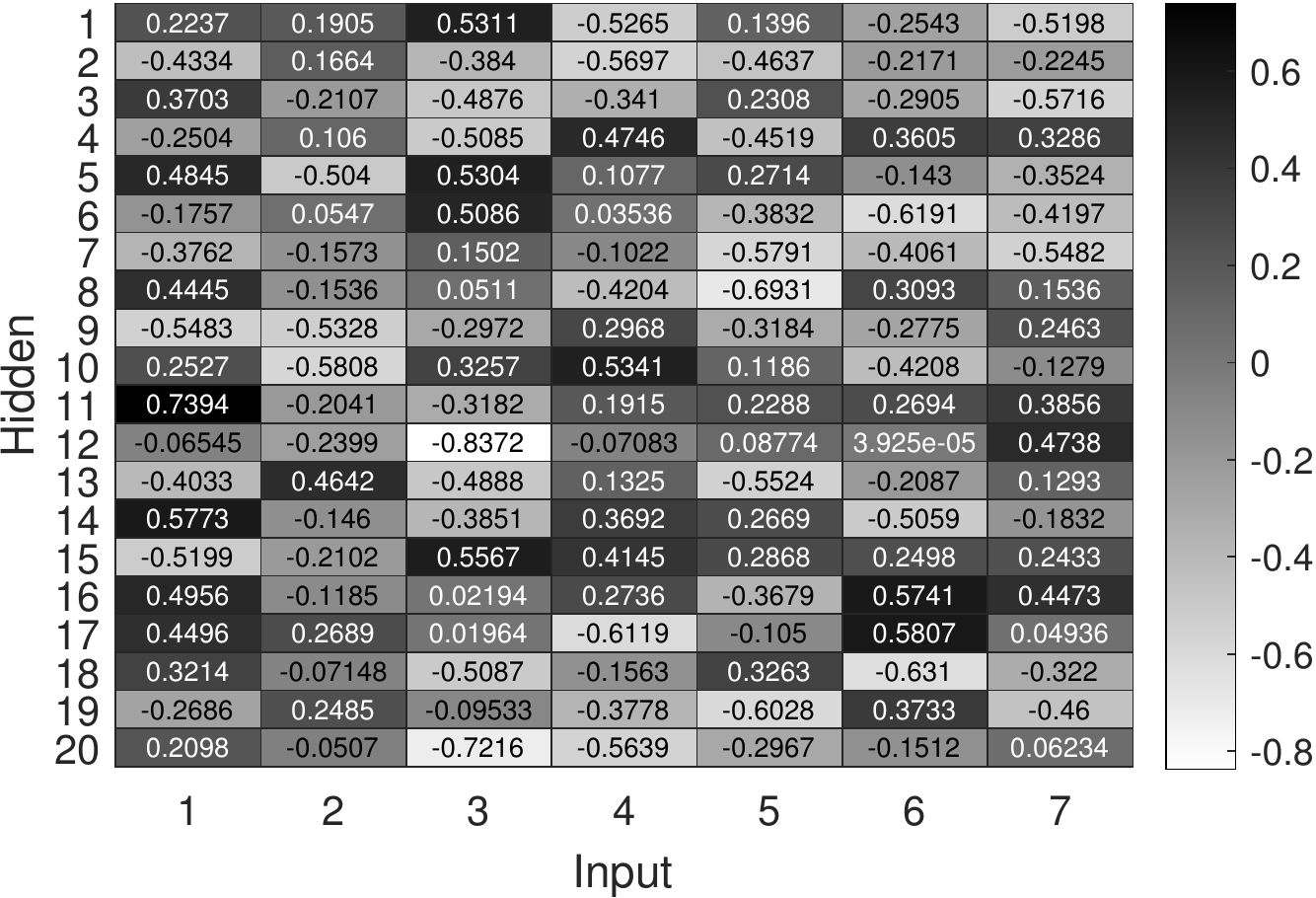}
\label{fig:hiddenWeights1}}
\subfloat[Connection weights after Summer1.]
{\includegraphics[width=0.5\columnwidth]{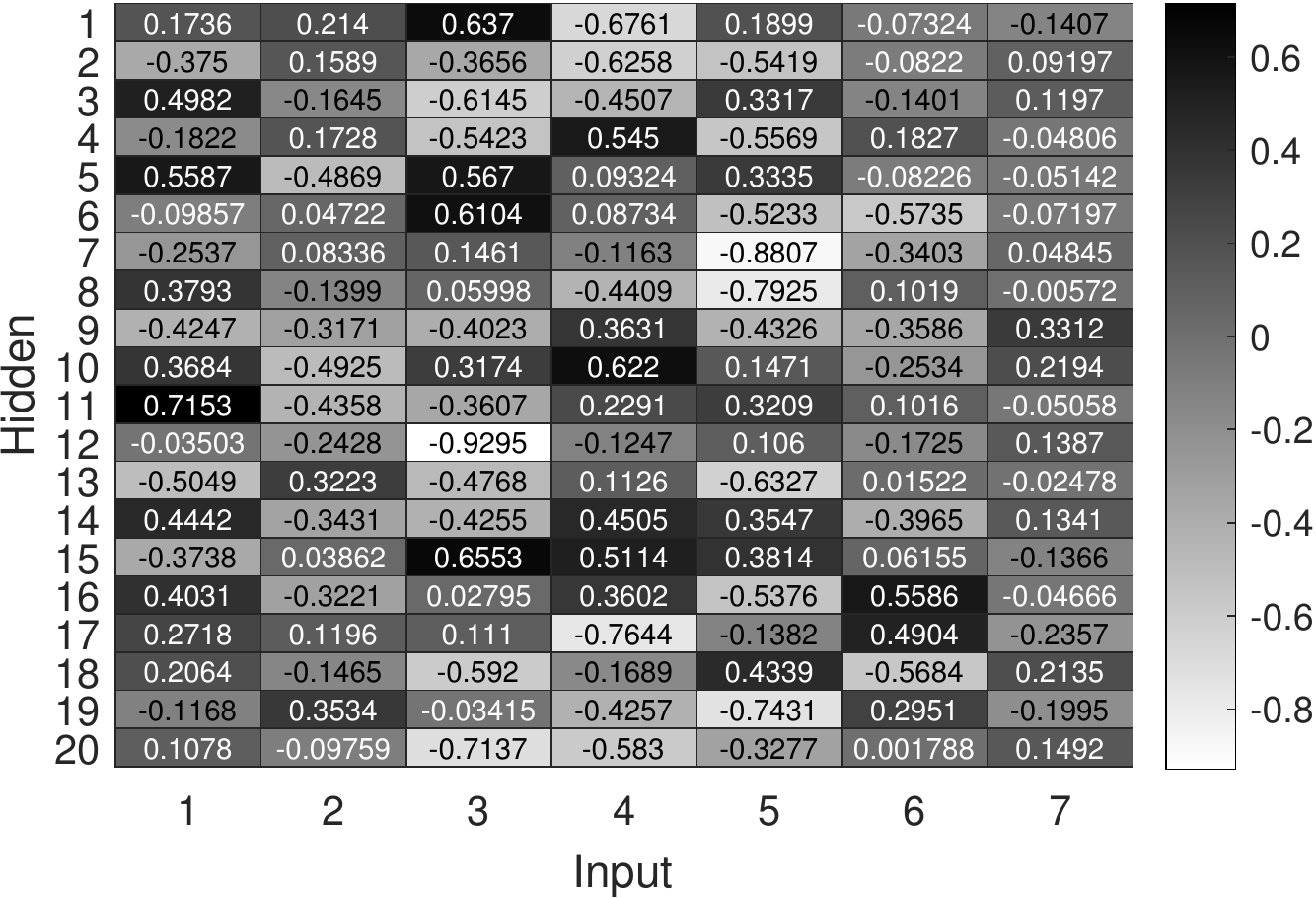}
\label{fig:hiddenWeights2}}

\subfloat[Connection weights after Winter1.]
{\includegraphics[width=0.5\columnwidth]{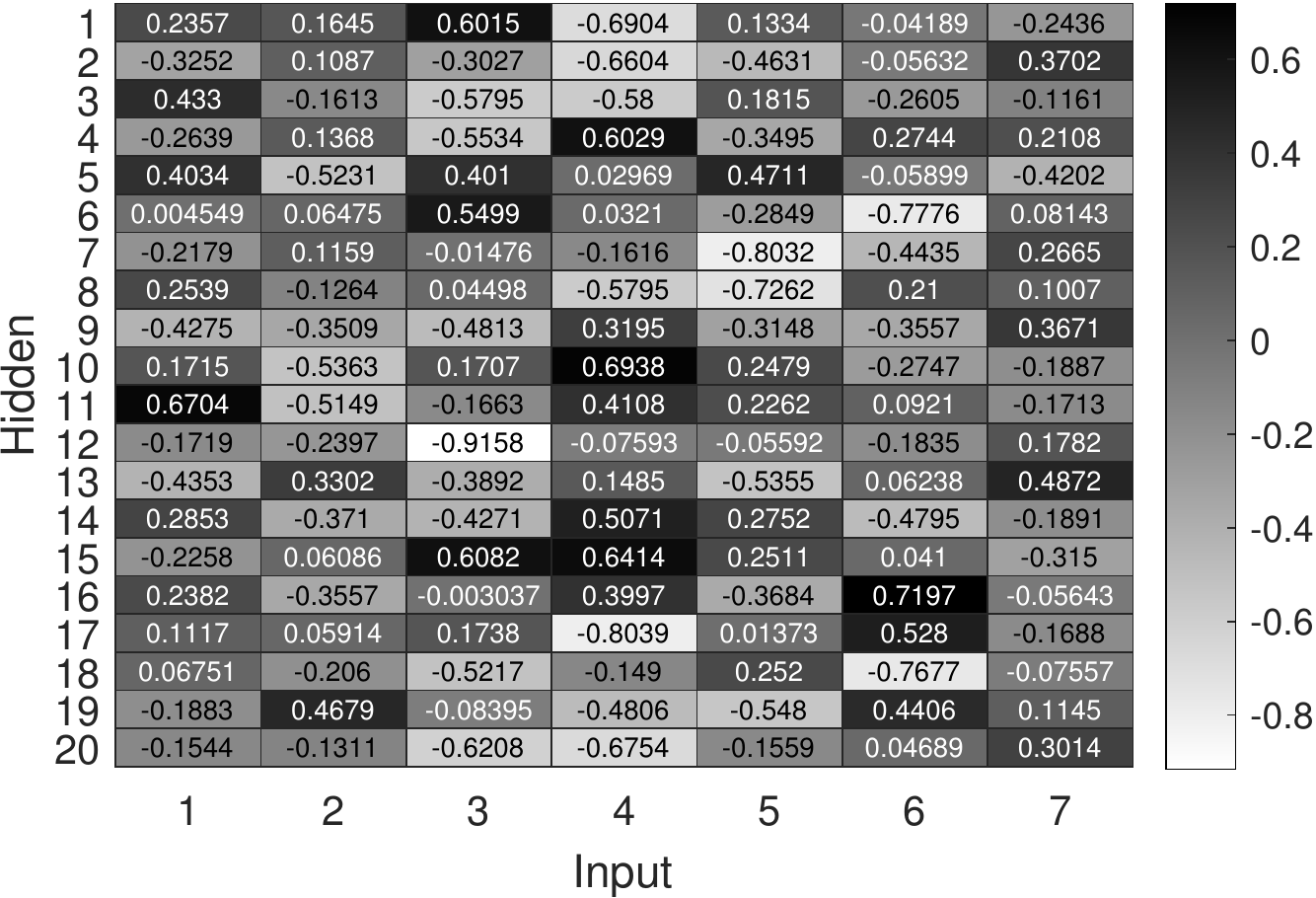}
\label{fig:hiddenWeights3}}
\subfloat[Connection weights after Summer2.]
{\includegraphics[width=0.5\columnwidth]{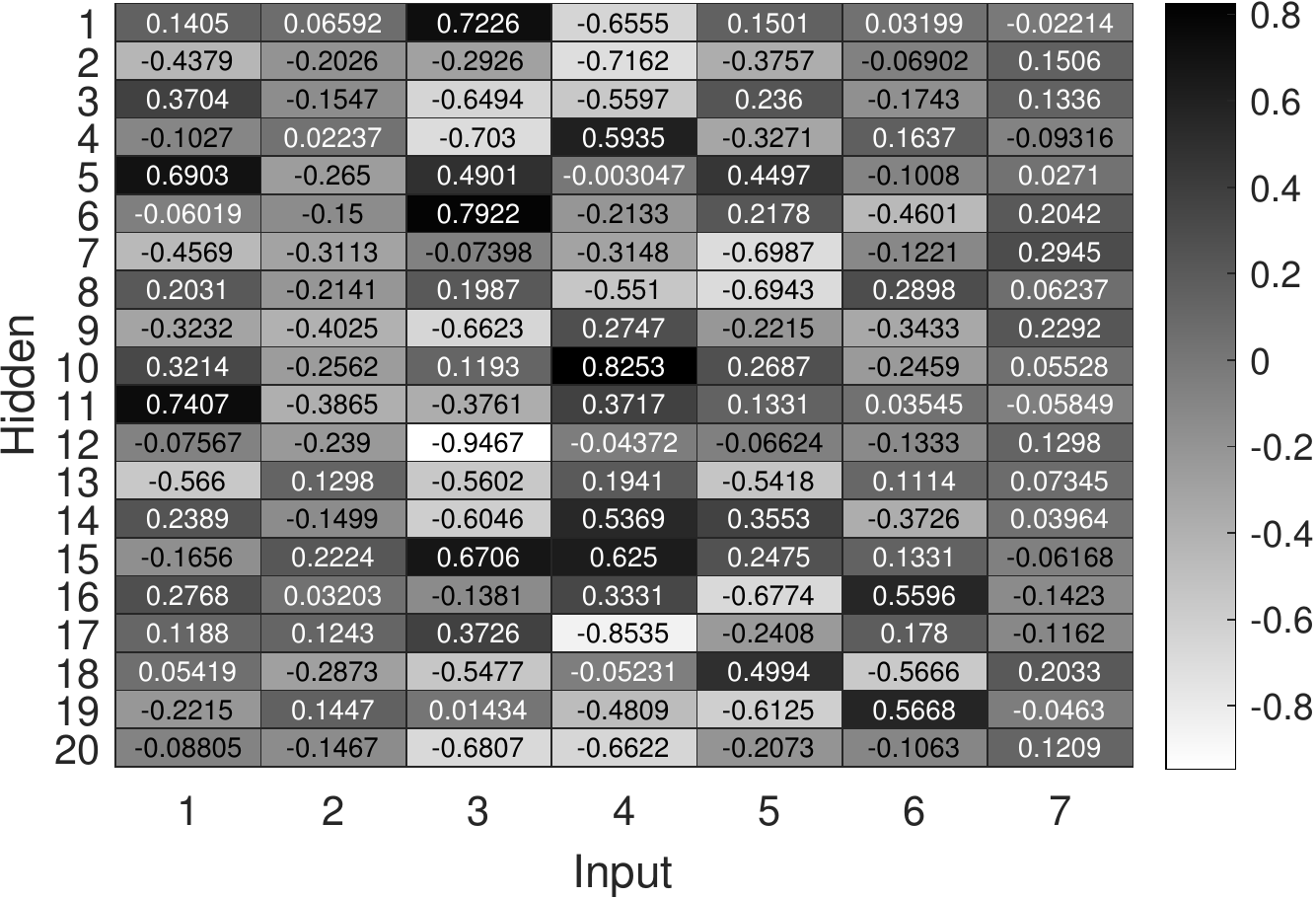}
\label{fig:hiddenWeights4}}

\end{subfigures}
\caption{Foraging task: heat map of the connection weights between the input and hidden neurons during a single run using the best ESP rule (ID:1). The $x$ and $y$-axes show the input and hidden neuron indices respectively (the 7th column shows the biases). Each connection on the heat map is color based on its value.}

\label{fig:hiddenWeights}
\end{figure*}

Figures \ref{fig:hiddenWeights} and \ref{fig:outputWeights} show the actual values of the connection weights between input and hidden, and between hidden and output layers, in a matrix form. Each column and row in the figures corresponds to a neuron in the input, hidden and output layers. In particular, Figures~\ref{fig:hiddenWeights1} and \ref{fig:outputWeights1} show the initial connection weights between input and hidden, and between hidden and output layers, sampled randomly. We performed three seasonal changes in the following order: summer, winter, summer. We used the best performing ESP rule (ID:1) to perform synaptic changes during the seasons. We provide the rest of the figures to show the connection weights after each consecutive season.

The connection weights between input and hidden layers appear to be distributed in the range $[-1,1]$; on the other hand, the connection weights between hidden and output layers tend to be distributed within the range $[0,1]$ at the end of each season. This may be due to the activation function of the output neurons, where only the neuron with the maximum activation value is allowed to fire.

\begin{landscape}
\centering
\begin{figure*}[ht]
\begin{subfigures}
\subfloat[Randomly sampled initial connection weights.]{\includegraphics[width = 19.2cm,trim=0 .5cm 0 0, clip]{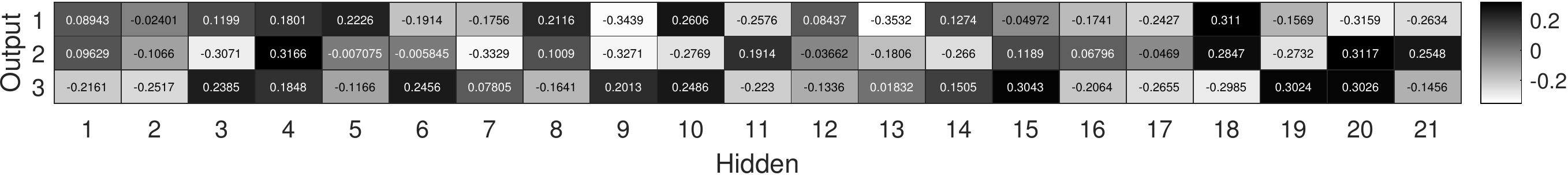}\label{fig:outputWeights1}}

\subfloat[Connection weights after Summer1.]{\includegraphics[width = 19.2cm,trim=0 .5cm 0 0, clip]{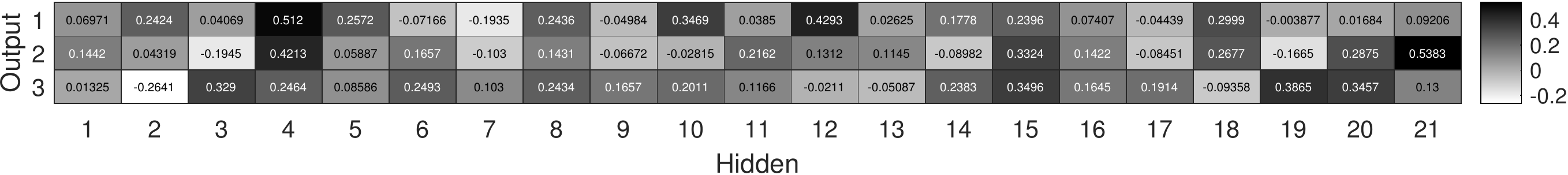}\label{fig:outputWeights2}}

\subfloat[Connection weights after Winter1.]{\includegraphics[width = 19.2cm,trim=0 .5cm 0 0, clip]{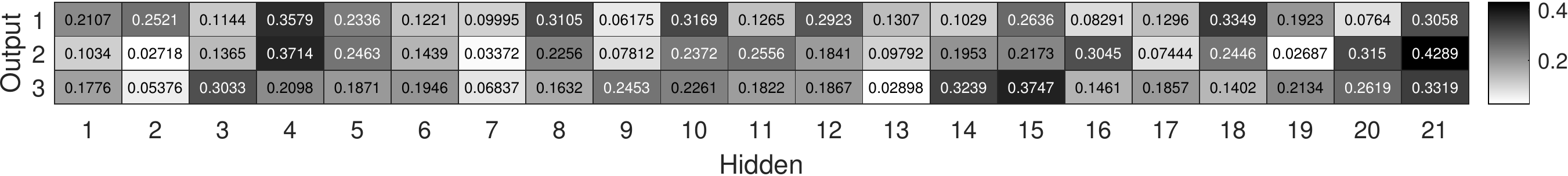}\label{fig:outputWeights3}}

\subfloat[Connection weights after Summer2.]{\includegraphics[width = 19.2cm]{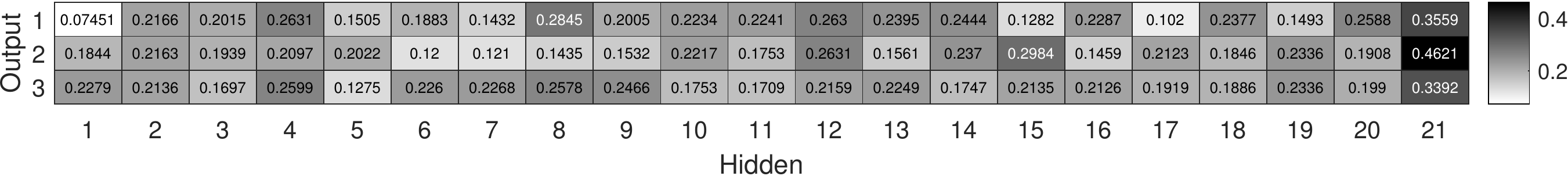}\label{fig:outputWeights4}}

\end{subfigures}
\caption{Foraging task: heat map of the connection weights between the hidden and output neurons during a single run using the best ESP rule (ID:1). The $x$ and $y$-axes show the hidden and output neuron indices respectively (the 21st column shows the biases). Each connection on the heat map is color coded based on its value.}
\label{fig:outputWeights}
\end{figure*}
\end{landscape}

\end{document}